%% file: neurips_2026.tex
\documentclass{article}

\PassOptionsToPackage{numbers, compress}{natbib}
\usepackage[main, final]{neurips_2026}

\usepackage[utf8]{inputenc}
\usepackage[T1]{fontenc}
\usepackage{url}
\usepackage{booktabs}
\usepackage{amsfonts}
\usepackage{amsmath,amssymb,amsthm,mathtools}
\usepackage{nicefrac}
\usepackage{microtype}
\usepackage{xcolor}
\usepackage{bm}
\usepackage{enumitem}
\usepackage{graphicx}
\usepackage{multirow}
\usepackage{algorithm}
\usepackage{algpseudocode}
\usepackage{hyperref}

\newtheorem{theorem}{Theorem}
\newtheorem{lemma}{Lemma}
\newtheorem{proposition}{Proposition}
\newtheorem{corollary}{Corollary}
\newtheorem{remark}{Remark}

\newcommand{\R}{\mathbb{R}}
\newcommand{\E}{\mathbb{E}}

\newcommand{\dtw}{d_{\mathrm{softDTW}}}
\newcommand{\norm}[1]{\left\lVert #1 \right\rVert}

\newcommand{\clip}[1]{\left[#1\right]_+}
\newcommand{\argmin}{\mathop{\mathrm{arg\,min}}}

\newcommand{\calC}{\mathcal{C}}
\newcommand{\calD}{\mathcal{D}}
\newcommand{\calZ}{\mathcal{Z}}

\title{Latent Barrier Steering:\\Hierarchical Safety for Generative Planning}

\author{%
  Renhao Zhang\\
  University of Massachusetts \\
  \texttt{renhaozhang@cs.umass.edu} \\
  \And
  Mingzhe Li\\
  University of Massachusetts \\
  \texttt{mingzheli@cs.umass.edu}\\
  \And
  Bruno Castro da Silva\\
  University of Massachusetts \\
  \texttt{bsilva@cs.umass.edu}
}

\begin{document}
\maketitle

\input{sec/0abstract.tex}
\input{sec/1introduction.tex}

\input{sec/2relatedworks.tex}
\input{sec/3preliminary.tex}
\input{sec/4methods.tex}

\input{sec/5experiment.tex}

\input{sec/6conclusion.tex}

\newpage
\begin{ack}
This work used computational resources and services provided by the Unity Research Computing Platform. We also thank the anonymous reviewers for their comments and suggestions, which helped strengthen and clarify this work.
\end{ack}

\bibliographystyle{plainnat}
\bibliography{refs}

\newpage
\appendix

\input{app/notation.tex}
\input{app/proofs.tex}
\input{app/algorithm.tex}
\input{app/domains.tex}
\input{app/ablations.tex}
\input{app/smoothness_experiments.tex}
\input{app/full_tables.tex}

\end{document}

%% file: sec/0abstract.tex
\begin{abstract}
We introduce \emph{Latent Barrier Steering} (LBS), a hierarchical semantic-to-path framework for safe flow-matching-based generative planning. Existing control-barrier-function (CBF) safety filters for diffusion and flow-matching planners act locally on generated samples, either during sampling or after prediction. Such repair is effective for small violations, but can struggle when a new constraint blocks the whole selected behavior mode rather than merely perturbing it locally. LBS decouples behavior selection from path certification through two coordinated interventions: it first steers the generator's behavior latent toward rollouts with larger safety margin, then decodes the steered latent and applies a path-space CBF quadratic program (QP) only for residual violations. We show that latent steering increases the margin available to the path-space corrector, yielding finite-time recovery guarantees and reduced correction burden. Experiments across navigation and robot end-effector planning tasks show that LBS preserves safety while improving goal reaching and reducing correction burden compared with path-level safety filters.
\end{abstract}

%% file: sec/1introduction.tex
\section{Introduction}
\label{sec:intro}

Robotic path planning must combine expressive behavior generation with strong safety guarantees. 
Generative policies, including diffusion and flow-matching planners, are attractive because they model multimodal trajectory distributions and synthesize diverse high-dimensional behaviors from demonstrations~\citep{ho2020ddpm,dhariwal2021diffusionbeatsgans,song2021scorebased,lipman2023flowmatching,janner2022diffuser,chi2023diffusionpolicy,carvalho2023motionplanningdiffusion,braun2024flowmatchingmotion}. 
Yet deployment safety is not inherited from offline training: new obstacles, workspace limits, or feasibility constraints can shift the test-time problem away from the demonstrations used to train the generator. 
This has motivated safe generative planners that combine learned samplers with control barrier functions (CBFs), which encode safety through a barrier-defined set, and impose constraints through quadratic programs or related filtering steps~\citep{ames2017cbf,glotfelter2017nonsmooth,nguyen2016exponential,xiao2019control,xiao2025safediffuser,dai2025safeflow,yang2025safeflowmatcher}. 
Existing CBF-based generative planners instantiate this idea primarily as path-level repair: they inject barrier constraints into the denoising or flow-generation process, or apply a CBF correction phase after a trajectory has been generated~\citep{xiao2025safediffuser,dai2025safeflow,yang2025safeflowmatcher}. 
In both cases, the intervention acts on sample or waypoint coordinates, pushing unsafe portions of the trajectory back toward the safe set while staying close to the generator's original output. 
This is effective when the generated trajectory is already nearly safe, so that only a short segment must be moved away from an obstacle. 
Under larger constraint shifts, however, the new constraint may block a substantial portion of the selected behavior mode. 
The corrector is then no longer making a local geometric repair: it must reshape a long connected segment of the rollout, which can increase intervention energy and leave the final trajectory safe but far from the goal.

\begin{figure}[t]
\centering
\includegraphics[width=\linewidth]{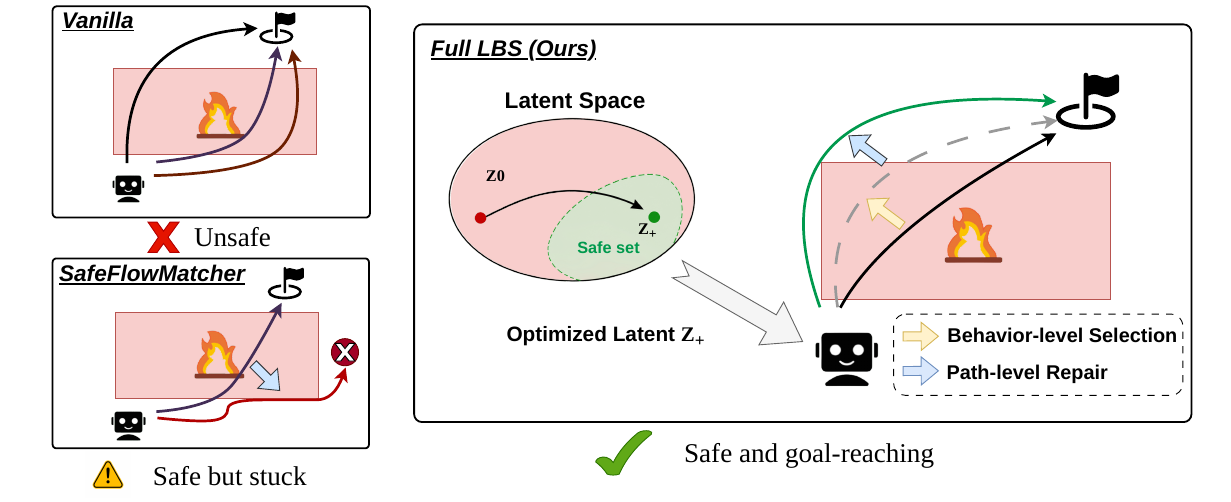}
\caption{
\textbf{Latent Barrier Steering separates behavior selection from path repair.}
A vanilla flow-matching planner may generate a goal-reaching but unsafe rollout under a new constraint. 
A path-space CBF corrector such as \textsc{SafeFlowMatcher} can enforce safety locally, but may produce a trajectory that is safe while no longer reaching the goal when the selected behavior mode is blocked. 
LBS first steers the behavior latent from $z_0$ to $z_+$ toward a safer rollout, then applies path-level CBF repair only for residual violations, yielding a trajectory that is both safe and goal-reaching.
}
\label{fig:scene_overview}
\vspace{-5mm}
\end{figure}

When path-level repair has to reshape a long segment of a rollout, the desired solution may be another behavior that the generator already knows how to produce. 
A multimodal demonstration dataset can contain several valid ways to solve the same start-goal task, such as passing above, below, or between obstacles, and generative policies can represent these alternatives as diverse rollouts. 
In standard generative planners, however, this diversity is usually accessed through stochastic sampling: the initial noise or sampling seed determines which behavior appears, with no direct handle for a safety filter to choose a rollout that is easier to certify. 
Geometry-aligned behavior latents, as in parameterized diffusion policies (PDP), provide such a handle by learning a low-dimensional coordinate in which nearby latent values decode to similar trajectories and larger latent moves induce more substantial behavior changes~\citep{zhang2026pdp}. 
This suggests a different division of labor for safe generative planning as shown in Figure~\ref{fig:scene_overview}: use the behavior latent to select or interpolate a safer rollout before asking a path-level CBF corrector to enforce the final constraints.

In this paper, we propose \emph{Latent Barrier Steering} (LBS), a two-stage framework for safe parameterized flow matching that separates behavior selection from path certification.
LBS first applies a CBF quadratic program (CBF-QP) in the behavior latent space to steer the decoded rollout toward a larger robust safety margin.
It then decodes the steered latent and applies a path-space CBF-QP only to the remaining waypoint-level violations.
This semantic-to-path structure lets the planner change the selected behavior before invoking local path repair. 
Theoretically, we show that latent steering improves the safety margin handed to the path-space corrector, yielding finite-time latent recovery when a safe latent is reachable and tighter path-level correction-time and intervention-energy certificates. 
Empirically, LBS improves safety, goal reaching, and correction efficiency across constrained planning tasks.
Our contributions are:
\begin{itemize}[leftmargin=1.5em,topsep=2pt,itemsep=1pt]
\item \textbf{Introduce semantic-to-path safety with parameterized flow matching.}
We build on a parameterized flow-matching planner whose geometry-aligned latent indexes different rollout behaviors. 
LBS uses this latent as the first point of safety intervention, steering the selected behavior before applying path-space CBF certification.

\item \textbf{Analyze when latent steering reduces path-space correction.}
We show that increasing the latent safety margin gives the path-space corrector a less-violating warm start. 
This yields finite-time latent recovery when a safe latent is reachable, and tighter correction-time and intervention-energy certificates for the downstream CBF-QP.

\item \textbf{Demonstrate safer and more goal-directed constrained planning.}
Across constrained planning tasks, LBS preserves safety while improving goal reaching and reducing correction energy compared with path-level safety filters. 
\end{itemize}

%% file: sec/2relatedworks.tex
\section{Related Work}
\label{sec:related}

\paragraph{Generative policies for planning.}
Diffusion and flow-based generative models provide expressive trajectory priors for robot planning and control, with diffusion planners modeling flexible multimodal behavior through iterative denoising and flow-matching planners improving sampling efficiency through deterministic transport \citep{ho2020ddpm,dhariwal2021diffusionbeatsgans,song2021scorebased,lipman2023flowmatching,liu2023rectified,janner2022diffuser,chi2023diffusionpolicy,wang2023offline,pearce2023imitating,carvalho2023motionplanningdiffusion,braun2024flowmatchingmotion}. 
These models have been applied to offline RL, visuomotor imitation, navigation, manipulation, and diverse human-demonstration datasets \citep{mandlekar2021robomimic,shafiullah2022bet,jia2024d3il}. 
Their strength is that a single learned sampler can represent many plausible trajectories for the same task context. 
However, the generated behavior is usually selected by conditioning variables or stochastic noise, so multimodal diversity does not by itself provide a controllable safety mechanism under new constraints.

\paragraph{Safety certification for generative planners.}
Control barrier functions (CBFs) and related invariance tools provide a standard way to enforce safety by maintaining the forward invariance of a superlevel set, often through a quadratic program that minimally modifies a nominal command \citep{ames2017cbf,glotfelter2017nonsmooth,nguyen2016exponential,xiao2019control}. 
Recent safe generative planners attach this machinery to diffusion or flow sampling: \textsc{SafeDiffuser} embeds finite-time invariance into the diffusion process, \textsc{SafeFlow} introduces flow-matching barrier functions for safe trajectory generation, and \textsc{SafeFlowMatcher} decouples prediction from correction by applying a finite-time CBF certificate only to the final path-space correction phase \citep{xiao2025safediffuser,dai2025safeflow,yang2025safeflowmatcher}. 
These methods provide formal safety mechanisms at test time, but their interventions are still expressed in sample or waypoint coordinates. 
They therefore primarily address local path repair after the generator has already selected a behavior.

\paragraph{Behavior latents and controllable multimodality.}
Controllable multimodality has a long history in parameterized actions, options, and skills, where a policy selects both a discrete behavior and continuous parameters that adapt the behavior to a task instance \citep{sutton1999between,konidaris2009skill,dasilva2012parameterized,hausknecht2016deep,masson2016parameterized,queisser2018bootstrapping,dalal2021primitives,fu2023meta,zhang2024model,gupta2025learning,zhang2026evidence}. 
Related work on skill discovery and latent-action imitation learns reusable behavior abstractions from demonstrations or interaction data \citep{niekum2015learning,shafiullah2022bet,lee2024latentactions,qiao2025offline,wu2025discrete}. 
PDP brings this idea to diffusion policies by learning a continuous behavior latent whose geometry is aligned with trajectory similarity, turning noise-driven diversity into an optimizable behavior handle \citep{zhang2026pdp}. 
These threads motivate our use of a behavior latent as the semantic coordinate for safety-aware steering before path-level certification.

%% file: sec/3preliminary.tex
\section{Background}
\label{sec:background}

\subsection{Parameterized flow matching for trajectory planning}
\label{sec:bg_gen}

Let $\calD$ denote the waypoint space, and let $\tau=(\tau^0,\ldots,\tau^H)\in\calD^{H+1}$ denote a planning trajectory over horizon $H$ under context $c$.
Flow matching learns a time-dependent vector field $v_t(\cdot,c;\theta)$ whose ODE transports a simple prior $p_0$ to the demonstration distribution $q(\tau\mid c)$ \citep{lipman2023flowmatching,liu2023rectified}: 
$\frac{d}{dt}\psi_t(\tau_0,c)=v_t(\psi_t(\tau_0,c),c;\theta)$, with $\tau_0\sim p_0$ and $\psi_1(\tau_0,c)\sim q(\cdot\mid c)$.
Following conditional flow matching, a standard objective samples $\tau_0\sim p_0$, $\tau_1\sim q(\cdot\mid c)$, $t\sim\mathrm{Unif}[0,1]$, sets $\tau_t=(1-t)\tau_0+t\tau_1$, and regresses to the optimal-transport velocity:
\begin{equation}
    \mathcal{L}_{\mathrm{FM}}(\theta)
    = \E\big[\norm{v_t(\tau_t,c;\theta)-(\tau_1-\tau_0)}_2^2\big].
    \label{eq:fm_loss}
\end{equation}
To expose a controllable behavior coordinate, we follow parameterized diffusion policies and learn a trajectory encoder $q_\phi(z\mid\tau)$ for a low-dimensional latent $z\in\calZ\subset\R^{d_z}$ \citep{zhang2026pdp}: 
\begin{equation}
    q_\phi(z\mid\tau)=\mathcal{N}\!\big(\mu_\phi(\tau),\operatorname{diag}(\sigma_\phi^2(\tau))\big),
    \qquad \bar z(\tau)=\mu_\phi(\tau),
    \label{eq:latent_encoder}
\end{equation}
where $\bar z(\tau)$ is the deterministic code used for conditioning.
The generator produces $\tau(z;\xi)=G_\theta(c,z,\xi)$, with stochastic seed $\xi$ or deterministic decode $\hat\tau_\theta(z,c)$. 
With latent prior $p(z)$ and nonnegative weights $\lambda_{\mathrm{rec}},\lambda_{\mathrm{KL}},\lambda_{\mathrm{geo}}$, we write a generic parameterized flow-matching objective as
\begin{align}
    \mathcal{L}_{\text{PFM}}(\theta,\phi)
    ={}& \mathcal{L}_{\mathrm{FM}}(\theta;z,c)
    + \lambda_{\mathrm{rec}}\,\E_{\tau,z\sim q_\phi}\!\big[\norm{\tau-\hat\tau_\theta(z,c)}_2^2\big]\nonumber\\
    &+ \lambda_{\mathrm{KL}}\,\E_{\tau}\!\big[D_{\mathrm{KL}}(q_\phi(z\mid\tau)\,\|\,p(z))\big]
    + \lambda_{\mathrm{geo}}\,\mathcal{L}_{\mathrm{geo}},
    \label{eq:pdp_objective}
\end{align}
where the geometry loss aligns trajectory and latent distances. 
For a sampled pair $(\tau_i,\tau_j)$, PDP computes a trajectory distance $\delta^X_{ij}=\dtw(X(\tau_i),X(\tau_j))$ between behavior-trace feature sequences $X(\tau_i)$ and $X(\tau_j)$ using soft-DTW \citep{cuturi2017softdtw}, and a latent distance $\delta^z_{ij}=\norm{\bar z(\tau_i)-\bar z(\tau_j)}_2$. 
It then penalizes $\mathcal{L}_{\mathrm{geo}}=\E_{(i,j)}[(\delta^z_{ij}-\kappa_{\mathrm{geo}}\delta^X_{ij})^2]$, with scale $\kappa_{\mathrm{geo}}>0$. 
This geometry term encourages $z$ to act as a behavior coordinate: nearby latents decode to similar
trajectory traces, while larger latent displacements correspond to larger behavioral changes.

\subsection{Control barrier functions for path safety}
\label{sec:bg_cbf}

For a control-affine system $\dot x=f(x)+g(x)u$ with state $x\in\calD$ and input $u\in\mathcal{U}$, and a $C^1$ barrier $b:\calD\to\R$, a control barrier function defines safety as $\calC=\{x:b(x)\ge 0\}$ and enforces forward invariance through the inequality $L_f b(x)+L_g b(x)u+\alpha(b(x))\ge 0$, where $L_f,L_g$ are Lie derivatives and $\alpha$ is a class-$\mathcal{K}$ rate \citep{ames2017cbf,glotfelter2017nonsmooth}. 
Given a nominal input $u^{\mathrm{nom}}$, the standard CBF-QP chooses the closest safe input:
\begin{equation}
    u^\star = \argmin_{u\in\mathcal{U}}\norm{u-u^{\mathrm{nom}}}_2^2\quad\text{s.t.}\quad L_f b(x)+L_g b(x)u+\alpha(b(x))\ge 0
    \label{eq:classic_cbf_qp}
\end{equation}
For states that may start outside $\calC$, fractional-power rates $\alpha(b)=\kappa_{\mathrm{cbf}}\,\mathrm{sgn}(b)|b|^\rho$ with rate constant $\kappa_{\mathrm{cbf}}>0$ and fractional-power exponent $0<\rho<1$ give finite-time convergence into $\calC$ \citep{nguyen2016exponential,xiao2019control,xiao2025safediffuser}.

Safe generative planners apply such constraints during diffusion or flow sampling, or as a correction phase after unconstrained prediction \citep{xiao2025safediffuser,dai2025safeflow,yang2025safeflowmatcher}. 
\textsc{SafeFlowMatcher}'s hard-correction phase satisfies, at sampling time $t$, for corrected waypoint $\tau_t^{c,k}$ (waypoint $k$ of the path at time $t$) and correction input $u_t^k$,
\begin{equation}
    \nabla b(\tau_t^{c,k})^\top u_t^k + \epsilon\,\mathrm{sgn}(b(\tau_t^{c,k})-\delta)\,|b(\tau_t^{c,k})-\delta|^\rho + w_t^k r_t^k \ge 0,
    \label{eq:sfm_certificate}
\end{equation}
where $\epsilon>0$ is the convergence rate, $0<\rho<1$ is the same fractional-power exponent as in \eqref{eq:classic_cbf_qp}, $\delta\ge 0$ is the robustness margin, and $w_t^k\ge 0$ is a vanishing weight schedule paired with a per-waypoint slack variable $r_t^k$ that together deactivate after switch time $t_w$.
The corrector enters $\calC_\delta=\{b\ge\delta\}$ within correction time
\begin{equation}
    T_{\mathrm{SFM}} \le t_w + \frac{(\delta-b(\tau_{t_w}^{c,k}))^{1-\rho}}{\epsilon(1-\rho)}.
    \label{eq:sfm_bound}
\end{equation}
This bound depends on the pre-correction barrier margin, the downstream quantity tightened in Section~\ref{sec:method}.

%% file: sec/4methods.tex
\section{Latent Barrier Steering}
\label{sec:method}

We introduce Latent Barrier Steering (LBS), a two-stage safety framework for generative planning that separates behavior-level steering from path-level certification. Prior CBF-based generative planners apply certified correction after a plain generator has already selected a behavior mode, leaving the corrector to repair the resulting rollout in waypoint or sample space. LBS instead builds on a parameterized flow-matching planner whose geometry-aligned behavior latent indexes different task-solving rollouts. At test time, Stage I treats this latent as the first safety intervention point: given a differentiable barrier function for the deployment constraint, it steers the latent toward a decoded rollout with a larger worst-case barrier margin. Stage II then uses this decoded rollout as a safety-aware warm start for a downstream path-space CBF-QP corrector, which enforces the remaining waypoint constraints under standard barrier-certificate conditions.

\subsection{LBS Objective with a Geometry-Aligned Latent}
\label{sec:latent_objective}

Stage I steers the behavior coordinate of a parameterized flow-matching planner. 
For a fixed task context $c$, each latent $z\in\mathcal Z$ decodes to a rollout
\[
    \tau(z)=G_\theta(c,z)
    =
    \bigl(\tau^0(z),\ldots,\tau^H(z)\bigr)
    \in \mathcal D^{H+1}.
\]
We use a deterministic decode $\tau(z)$ by fixing the sampling randomness. 
The geometry-regularized objective in Section~\ref{sec:bg_gen} encourages local latent moves to induce controlled rollout changes. 
Because the encoder is KL-regularized toward the prior $p(z)=\mathcal N(0,I)$, Stage~I restricts steering to a compact region $\Omega\subseteq\mathcal Z$ around the prior center, implemented by latent clipping. 
The prior is used only to organize the latent space during training. 
At test time, Stage~I optimizes a single latent within $\Omega$, which contains the encoded demonstration modes and their interpolation region. 
On this region, we assume the decoder and barrier are locally Lipschitz. 
This is an analysis assumption encouraged by the geometry objective rather than guaranteed by it (Appendix~\ref{app:geom_remark}). 
Concretely, for $z,z'\in\Omega$ and $x,x'\in\mathcal D$,
\[
    \|\tau(z)-\tau(z')\| \le L_\tau \|z-z'\|,
    \qquad
    |b(x)-b(x')|\le L_b\|x-x'\|.
    \label{eq:latent_path_lipschitz}
\]
This local geometry is the condition that makes latent-space safety steering meaningful: a controlled update in $z$ produces a controlled change in the decoded rollout.

We now define the latent safety objective. 
Using the barrier $b$ and robustness margin $\delta$ from Section~\ref{sec:bg_cbf}, Stage~I assigns each latent a differentiable approximation to the worst shifted waypoint margin of its decoded rollout:
\[
    m_\beta(z)
    :=
    -\frac{1}{\beta}
    \log
    \sum_{k=0}^{H}
    \exp\!\left(
        -\beta\bigl(b(\tau^k(z))-\delta\bigr)
    \right),
    \qquad \beta>0 .
    \label{eq:softmin_margin}
\]
This soft-min margin satisfies
\[
    \min_k\bigl(b(\tau^k(z))-\delta\bigr)
    -
    \frac{\log(H+1)}{\beta}
    \le
    m_\beta(z)
    \le
    \min_k\bigl(b(\tau^k(z))-\delta\bigr),
    \label{eq:softmin_bound}
\]
so $\varepsilon_{\mathrm{soft}}=\log(H+1)/\beta$ is the only gap between $m_\beta$ and the true worst-waypoint margin; see Appendix~\ref{app:eq_softmin_bound}.
Combining this construction with the local geometry above gives
\[
    |m_\beta(z)-m_\beta(z')|
    \le
    L_bL_\tau\|z-z'\|,
    \qquad z,z'\in\Omega ,
    \label{eq:margin_lipschitz}
\]
with proof in Appendix~\ref{app:eq_margin_lipschitz}. 
Thus, controlled motion in the behavior latent produces controlled changes in the rollout safety margin. 
The strongest Stage~I target is the safe-latent set $\mathcal Z_\delta:=\{z\in\mathcal Z:m_\beta(z)\ge0\}$, whose elements decode to rollouts satisfying the buffered waypoint constraint up to the soft-min gap. 
LBS, however, does not require Stage~I to reach this set before handoff: increasing $m_\beta$ already reduces the barrier deficit that Stage~II must correct. 
This is the margin quantity used in Section~\ref{sec:stage1} for latent CBF steering and in Section~\ref{sec:stage2} to relate latent progress to the path-space correction burden. 
Because $m_\beta$ is defined over discrete waypoints, all guarantees following are stated at the waypoint level. Appendix~\ref{app:eq_segment_constant} extends these margins to continuous path segments.

\subsection{Stage I: Latent CBF Steering}
\label{sec:stage1}

Stage~I uses the latent margin $m_\beta$ from Section~\ref{sec:latent_objective} as a barrier over the behavior latent. 
Rather than modifying waypoints independently, it updates $z$ so that the pretrained decoder produces a rollout with larger safety margin before path-space correction. 
We model latent steering as $\dot z=u_z$ and choose the smallest modification to a reference latent update $u_z^{\mathrm{ref}}$. 
In our implementation, $u_z^{\mathrm{ref}}(z)=-\gamma_z z$ with $\gamma_z>0$, which mildly contracts the latent toward the prior center and discourages off-support drift. 
At each latent $z$, LBS solves
\begin{equation}
    u_z^\star(z)
    =
    \arg\min_{u\in\mathbb R^{d_z}}
    \|u-u_z^{\mathrm{ref}}(z)\|_2^2
    \quad
    \mathrm{s.t.}\quad
    \nabla m_\beta(z)^\top u
    +
    \kappa\,\operatorname{sgn}(m_\beta(z))|m_\beta(z)|^\rho
    \ge 0 ,
    \label{eq:latent_cbf_qp}
\end{equation}
where $\kappa>0$ and $0<\rho<1$ are the CBF rate and exponent. 
Because \eqref{eq:latent_cbf_qp} has one affine constraint, it is the projection of $u_z^{\mathrm{ref}}$ onto a half-space. 
Let $\psi(z):=\nabla m_\beta(z)^\top u_z^{\mathrm{ref}}(z)+\kappa\,\operatorname{sgn}(m_\beta(z))|m_\beta(z)|^\rho$. 
Then, whenever $\nabla m_\beta(z)\ne0$,
\begin{equation}
    u_z^\star(z)
    =
    u_z^{\mathrm{ref}}(z)
    +
    \frac{[-\psi(z)]_+}{\|\nabla m_\beta(z)\|_2^2}\nabla m_\beta(z).
    \label{eq:latent_projection}
\end{equation}
Thus, Stage~I follows the reference update when it already satisfies the latent barrier condition, and otherwise adds the minimum correction along $\nabla m_\beta$; the derivation is in Appendix~\ref{app:eq_closed_form}.

\begin{proposition}[Latent margin decrease]
\label{prop:latent_descent}
Let $V(z):=[-m_\beta(z)]_+$. 
Along the Stage~I flow $\dot z=u_z^\star(z)$, whenever $m_\beta(z)<0$ and $\nabla m_\beta(z)\ne0$,
\[
    \dot V(z(t))\le -\kappa V(z(t))^\rho .
\]
\end{proposition}

Proposition~\ref{prop:latent_descent} says that Stage~I decreases the margin deficit whenever the unsafe decoded rollout has a first-order latent direction for improvement. 
The proof is a direct consequence of the CBF constraint and is given in Appendix~\ref{app:eq_V_dot}. 
When this first-order sensitivity is uniformly nonzero on the unsafe part of $\Omega$, the same fractional-power rate gives finite-time recovery.

\begin{theorem}[Finite-time Stage-I recovery]
\label{thm:latent_ft}
Suppose the latent flow $\dot z=u_z^\star(z)$ remains in $\Omega$ before reaching $\mathcal Z_\delta$, and suppose there exists $g_m>0$ such that $\|\nabla m_\beta(z)\|_2\ge g_m$ for all $z\in\Omega$ with $m_\beta(z)<0$. 
Then, for any $z_0\in\Omega$ with $m_\beta(z_0)<0$, $z(t)\in\mathcal Z_\delta$ for all $t\ge T_z:=|m_\beta(z_0)|^{1-\rho}/(\kappa(1-\rho))$. 
Moreover, $\mathcal Z_\delta$ is forward-invariant.
\end{theorem}

The proof is in Appendix~\ref{app:proof_latent_ft}. 
In practice, Stage~I uses projected Euler steps with either backpropagated or finite-difference estimates of $\nabla m_\beta$; implementation details and generator-call cost are given in Appendix~\ref{app:lbs_algorithm}.

\subsection{Stage II Correction and Handoff}
\label{sec:stage2}

Stage~I returns a steered latent $z_+$ and decoded rollout $\tau(z_+)$. 
Stage~II then applies the path-space CBF-QP from Section~\ref{sec:bg_cbf} to certify the remaining waypoint-level violations. 
The key question is whether improving the latent margin $m_\beta$ actually reduces the path-space violation.

Let $\tilde b_{t_w}(z):=\min_k(b(\tau_{t_w}^{c,k}(z))-\delta)$ denote the shifted path margin at the start of hard correction, where $t_w$ is the switch time used by the path-space corrector. 
For predictor--corrector methods, the switch-time path may differ from the decoded rollout; write this deviation as $\varepsilon_{\mathrm{pred}}(z):=\max_k\|\tau_{t_w}^{c,k}(z)-\tau^k(z)\|_2$. 
For a post-hoc corrector applied directly to $\tau(z)$, $\varepsilon_{\mathrm{pred}}(z)=0$.

\begin{proposition}[Latent-to-path margin transfer]
\label{prop:transfer}
Under the local regularity condition in Section~\ref{sec:latent_objective}, for all $z\in\Omega$,
\[
    |\tilde b_{t_w}(z)-m_\beta(z)|
    \le
    \varepsilon_{\mathrm{soft}}+L_b\varepsilon_{\mathrm{pred}}(z),
    \qquad
    \varepsilon_{\mathrm{soft}}=\frac{\log(H+1)}{\beta}.
\]
\end{proposition}

Proposition~\ref{prop:transfer} is the formal handoff from latent steering to path correction: increasing $m_\beta(z)$ improves the initial path-space margin up to the soft-min gap and any predictor deviation. 
The proof is in Appendix~\ref{app:proof_transfer}. 
Thus, Stage~I does not need to certify the trajectory by itself; it only needs to reduce the deficit that Stage~II starts from.

\begin{corollary}[Reduced Stage-II correction burden]
\label{cor:stage2_burden}
Let $V_+:=[-\tilde b_{t_w}(z_+)]_+$ be the initial Stage-II barrier deficit. 
With Stage-II CBF rate $\epsilon>0$ and exponent $\rho\in(0,1)$, the path-space corrector satisfies
\[
    T_{\mathrm{corr}}(z_+)
    \le
    t_w+\frac{V_+^{1-\rho}}{\epsilon(1-\rho)}.
\]
If the squared intervention magnitude is bounded by $c_{II}V_{\mathrm{path}}^{2\rho}$ during correction, then
\[
    J_{II}(z_+)
    \le
    \frac{c_{II}V_+^{1+\rho}}{\epsilon(1+\rho)}.
\]
\end{corollary}

Corollary~\ref{cor:stage2_burden} shows why the latent stage is useful even when it does not reach $\mathcal Z_\delta$: by decreasing $V_+$, Stage~I tightens both the certified correction-time bound and the intervention-energy bound. 
The proof follows the standard fractional-power CBF argument and is given in Appendix~\ref{app:proof_stageII_energy}.

It remains to choose when to stop latent steering. 
Running Stage~I longer reduces the terminal deficit but incurs additional latent-steering cost. 
In the exact-gradient limit, the descent law from Proposition~\ref{prop:latent_descent} gives $V_+^{1-\rho}=V_0^{1-\rho}-\eta\kappa(1-\rho)N_z$, where $V_0:=[-m_\beta(z_0)]_+$. 
Combining this decrease with a per-step Stage-I cost $c_I$ and the Stage-II energy bound above yields a cost-sensitive stopping rule.

\begin{theorem}[Cost-sensitive Stage-I handoff]
\label{thm:joint_energy}
Suppose the Stage-I decrease law holds along the latent trajectory and the margin-transfer slack in Proposition~\ref{prop:transfer} is absorbed into the Stage-II deficit. 
For $V_0>0$, the resulting upper bound on total cost is minimized at $V_+^\star=(c_I\epsilon/(\eta\kappa c_{II}))^{1/(2\rho)}$. 
The corresponding handoff iteration is $N_z^\star=(V_0^{1-\rho}-(V_+^\star)^{1-\rho})/(\eta\kappa(1-\rho))$ when $V_+^\star<V_0$; otherwise, LBS hands off to Stage~II immediately.
\end{theorem}

Theorem~\ref{thm:joint_energy} says that Stage~I should stop once another latent update is expected to cost more than the correction energy it saves downstream. 
The convexity argument and the explicit total-cost expression are deferred to Appendix~\ref{app:proof_joint_energy}.

%% file: sec/5experiment.tex
\section{Experiments}
\label{sec:experiments}

We evaluate whether combining behavior-latent steering with path-space CBF certification improves constrained flow-matching planning.
The experiments are organized around four questions.
\textbf{Q1:} How does LBS compare with existing safe generative planners across different domains and test-time safety constraints?
\textbf{Q2:} How does LBS change the generated behavior before path-space correction?
\textbf{Q3:} How effective is Stage~I's latent steering design compared with alternative ways of selecting the behavior latent?
\textbf{Q4:} What does Stage~I achieve on its own, and what does it hand off to Stage~II?

\begin{figure}[t]
    \centering
    \includegraphics[width=0.8\linewidth]{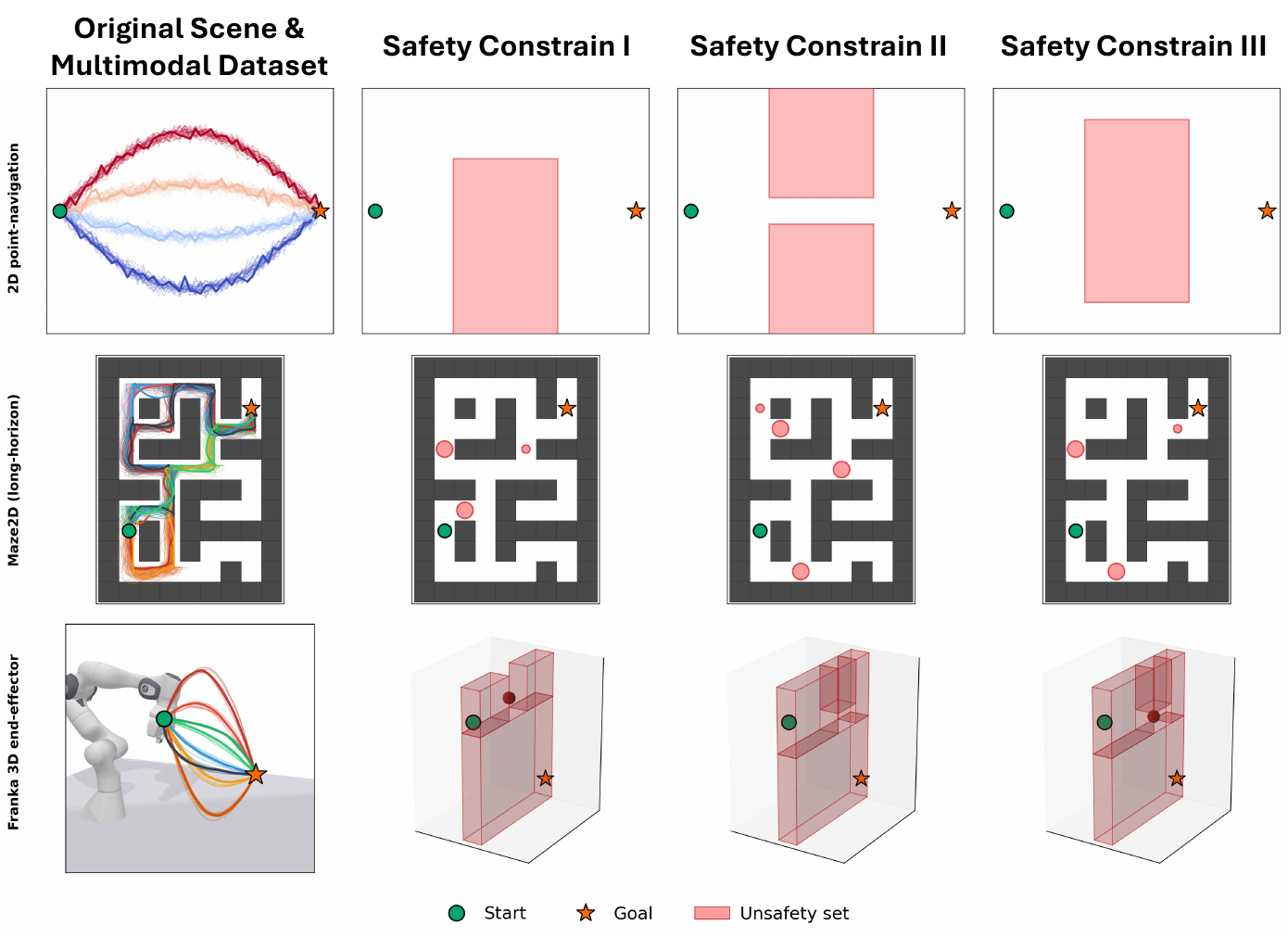}
    \caption{
    \textbf{Experimental domains and test-time safety constraints.}
    Each row shows one planning domain: 2D point-navigation, Maze2D, and Franka 3D end-effector planning. 
    The first column visualizes the original scene and multimodal demonstration dataset, where multiple behaviors can reach the same goal. 
    The remaining columns show three test-time safety constraints imposed on the same pretrained planner. 
    Green circles denote starts, orange stars denote goals, and red shaded regions denote unsafe sets. 
    These constraints test whether a method can preserve safety and goal reaching when certification requires more than local path repair.
    }
    \label{fig:domains}
\end{figure}

\paragraph{Setup.}
We evaluate on constrained planning tasks spanning low-dimensional navigation, long-horizon maze navigation, and robot end-effector planning. 
In each domain, the pretrained flow-matching planner is trained on a multimodal dataset with several feasible ways to reach the goal. 
At test time, we keep the pretrained planner fixed and introduce three different safety constraints, shown in Figure~\ref{fig:domains}. 
These constraints vary how much the originally generated rollout must change before it can be certified, but the method is not given this structure explicitly. 
Full domain construction, obstacle layouts, model architectures, and training details are provided in Appendix~\ref{app:domains}.

\paragraph{Baselines and Metrics.}
We compare LBS against vanilla flow-matching planner, \textsc{SafeDiffuser}~\citep{xiao2025safediffuser}, \textsc{SafeFlow}~\citep{dai2025safeflow}, and \textsc{SafeFlowMatcher}~\citep{yang2025safeflowmatcher}, which corresponds to using Stage~II without latent steering. 
We report Barrier Safety (referred to as Safety below, with Maze2D walls handled as described in Appendix~\ref{app:maze}), Reach, final goal distance (ED), CBF intervention Energy, curvature smoothness (CS), and per-episode Stage-I/Stage-II runtime. 
Together, these metrics distinguish whether a method is safe, goal-reaching, or both, with low correction burden.

\subsection{Performance Comparison Across Domains and Constraints}
\label{sec:q1_main}

Table~\ref{tab:q1_main_compact} reports Safety, Reach, and intervention Energy for all three constraints in each domain. Final goal distance, curvature smoothness, and runtime are reported in Appendix~\ref{app:full_results}.
This comparison evaluates whether LBS improves the overall safety--performance tradeoff relative to existing path-level safety filters under the same pretrained planner and test-time constraints.

Across domains and constraints, the table shows a clear separation between goal-directed generation and certified safety. 
The unconstrained flow-matching planner often preserves the task behavior learned from demonstrations, so its rollouts remain close to the goal, but it has no mechanism to adapt that behavior to new safety constraints. 
This leads to goal-reaching trajectories that are frequently unsafe. 
Existing CBF-based safety filters correct this failure in the opposite direction: they reliably improve safety, but because the intervention is applied in sample or waypoint space after a behavior has already been selected, the corrected trajectory can lose the structure needed to complete the task. 
This is reflected not only in reduced reach, but also in larger final distance (Table~\ref{tab:q1_main_all}) and higher correction energy.

\begin{table*}[t]
\centering
\caption{
\textbf{Performance comparison across domains and test-time safety constraints.}
Each domain uses the same pretrained flow-matching planner across methods. 
Final goal distance, curvature smoothness, and runtime are reported in Appendix~\ref{app:full_results}.
}
\label{tab:q1_main_compact}
\scriptsize
\setlength{\tabcolsep}{3.2pt}
\resizebox{\textwidth}{!}{%
\begin{tabular}{@{}llccccccccc@{}}
\toprule
& & \multicolumn{3}{c}{2D} & \multicolumn{3}{c}{Franka} & \multicolumn{3}{c}{Maze2D} \\
\cmidrule(lr){3-5}\cmidrule(lr){6-8}\cmidrule(lr){9-11}
Constraint & Method
& Safety$\uparrow$ & Reach$\uparrow$ & Energy$\downarrow$
& Safety$\uparrow$ & Reach$\uparrow$ & Energy$\downarrow$
& Safety$\uparrow$ & Reach$\uparrow$ & Energy$\downarrow$ \\
\midrule
\multirow{5}{*}{I}
& Vanilla FM               & 12\% & \textbf{100\%} & --    & 12\% & \textbf{100\%} & --     & 0\%  & 50\% & -- \\
& \textsc{SafeDiffuser}    & \textbf{100\%} & 10\% & 1.226 & \textbf{100\%} & 0\%  & 0.0023 & \textbf{100\%} & 0\% & 0.0288 \\
& \textsc{SafeFlow}        & \textbf{100\%} & 18\% & 1.077 & \textbf{100\%} & 16\% & 0.0020 & \textbf{100\%} & 0\% & 0.1208 \\
& \textsc{SafeFlowMatcher} & \textbf{100\%} & 18\% & 1.166 & \textbf{100\%} & 14\% & 0.0075 & \textbf{100\%} & 0\% & 0.2668 \\
& \textbf{Full LBS}        & \textbf{100\%} & \textbf{100\%} & \textbf{0.037} & \textbf{100\%} & \textbf{100\%} & \textbf{0.0004} & \textbf{100\%} & \textbf{95\%} & \textbf{0.010} \\
\midrule
\multirow{5}{*}{II}
& Vanilla FM               & 0\%  & \textbf{100\%} & --    & 0\%  & \textbf{100\%} & --     & 38\% & 50\% & -- \\
& \textsc{SafeDiffuser}    & \textbf{100\%} & 34\% & 0.728 & \textbf{100\%} & 0\% & 0.0023 & \textbf{100\%} & 7\%  & \textbf{0.0010} \\
& \textsc{SafeFlow}        & \textbf{100\%} & 20\% & 0.976 & \textbf{100\%} & 0\% & 0.0023 & \textbf{100\%} & 25\% & 0.0624 \\
& \textsc{SafeFlowMatcher} & \textbf{100\%} & 22\% & 1.051 & \textbf{100\%} & 0\% & 0.0086 & 98\%  & 18\% & 0.1404 \\
& \textbf{Full LBS}        & \textbf{100\%} & 90\% & \textbf{0.160} & \textbf{100\%} & \textbf{100\%} & \textbf{0.0004} & \textbf{100\%} & \textbf{90\%} & 0.009 \\
\midrule
\multirow{5}{*}{III}
& Vanilla FM               & 0\%  & \textbf{100\%} & --    & 0\%  & \textbf{100\%} & --     & 28\% & 50\% & -- \\
& \textsc{SafeDiffuser}    & \textbf{100\%} & 0\%  & 1.321 & \textbf{100\%} & 0\% & \textbf{0.0023} & \textbf{100\%} & 7\%  & 0.0072 \\
& \textsc{SafeFlow}        & \textbf{100\%} & 4\%  & 1.262 & \textbf{100\%} & 0\% & \textbf{0.0023} & \textbf{100\%} & 38\% & 0.0345 \\
& \textsc{SafeFlowMatcher} & \textbf{100\%} & 4\%  & 1.366 & \textbf{100\%} & 0\% & 0.0086 & \textbf{100\%} & 53\% & 0.0740 \\
& \textbf{Full LBS}        & \textbf{100\%} & 82\% & \textbf{0.303} & \textbf{100\%} & 98\% & 0.0028 & \textbf{100\%} & \textbf{93\%} & \textbf{0.004} \\
\bottomrule
\end{tabular}%
}
\end{table*}

Full LBS changes this pattern by intervening before path-space certification: the latent stage can change the selected behavior, giving Stage~II a rollout that is closer to both the safe set and the task solution. 
This explains why LBS preserves goal reaching while maintaining high safety, and why its CBF intervention energy is lower in most settings. 
Because all methods use the same pretrained planner within each domain, the gains come from where safety intervention is applied at test time rather than from a stronger generator or additional task-specific training. 
The runtime columns in Table~\ref{tab:q1_main_all} reflect the current implementation tradeoff: Stage~I adds overhead due to zeroth-order finite-difference gradients, while Stage~II remains comparable to path-level baselines.
This overhead grows linearly with the behavior-latent dimension $d_z$ (Appendix~\ref{app:stage_i_cost}), which is set by the number of demonstrated behavior modes rather than by the horizon or robot degrees of freedom ($d_z\le4$ in all domains), and the resulting $0.35$--$4.50$\,s of Stage~I is incurred once before execution.
Thus, under the present implementation, the main efficiency gain appears as reduced correction burden rather than lower wall-clock time; cheaper latent-gradient estimators should reduce the Stage~I cost while preserving the qualitative advantage.

\subsection{Ablation: Latent Steering Provides the Warm Start}
\label{sec:ablation_warm_start}

\begin{figure}[t]
    \centering
    \includegraphics[width=\linewidth]{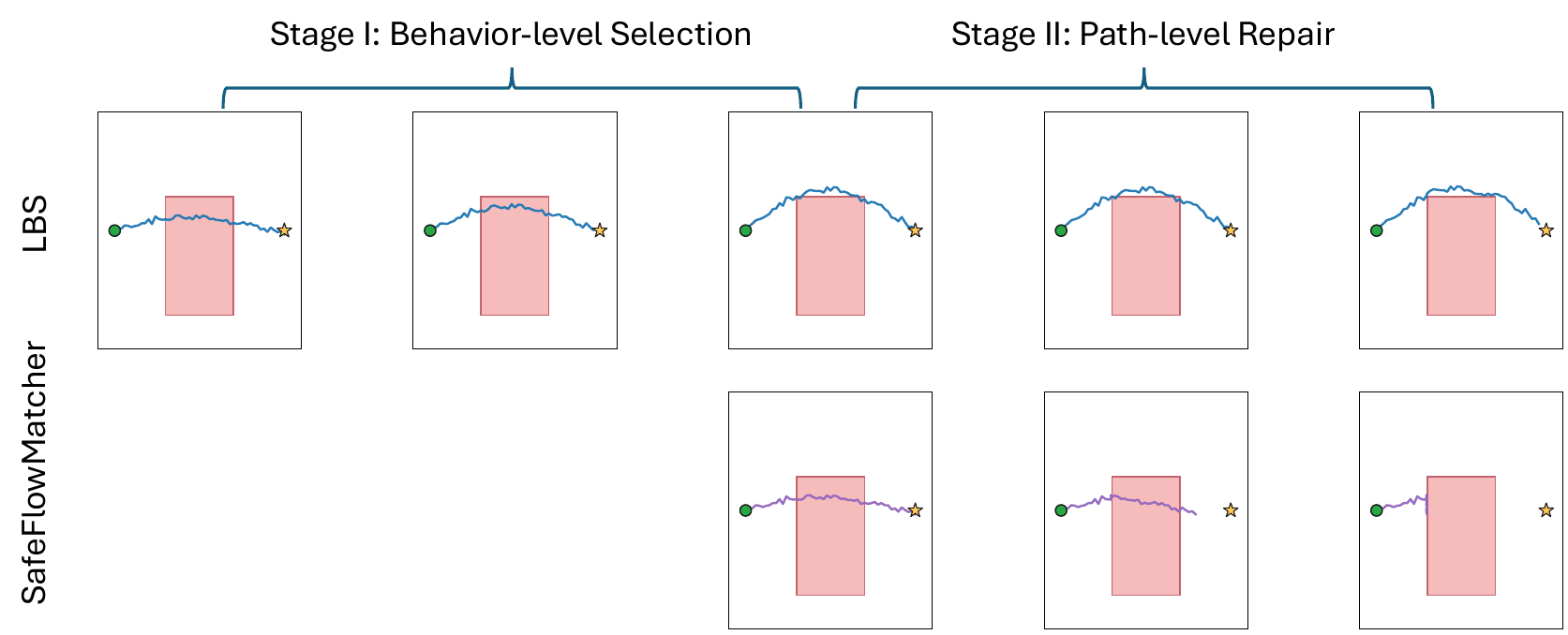}
    \caption{
    \textbf{Latent steering provides the warm start needed for path-space repair.}
    Stage~I performs behavior-level selection by steering the latent and changing the decoded rollout before certification; Stage~II then applies path-level CBF repair to remove residual violations. 
    The second row removes Stage~I and applies the same path-space correction directly, corresponding to \textsc{SafeFlowMatcher}. 
    Without the latent-steered warm start, the local corrector remains tied to the original rollout and fails to produce a goal-reaching safe path. 
    With Stage~I, the corrected trajectory starts from a better behavior and Stage~II only needs to repair the remaining local violation.
    }
    \label{fig:warm_start_ablation}
\end{figure}

The main comparison shows that LBS improves the safety--goal-reaching tradeoff, but the table alone does not explain how the two stages interact. 
We therefore visualize a representative episode in Figure~\ref{fig:warm_start_ablation}, comparing path-space correction with and without the latent-steering stage. 
The figure is organized as two rows with five columns. 
The first row shows Full LBS: the rollout is first decoded from the initial behavior latent, then updated through Stage~I, and finally passed to Stage~II for path-space CBF correction. 
The second row removes Stage~I and applies the same Stage~II corrector directly to the original decoded rollout. 
The first panel of Full LBS is therefore the same rollout as the Stage-II-only warm start shown in the third panel of the second row.

This visualization isolates the role of Stage~I. 
Before path-space correction, the initial rollout is unsafe and poorly positioned for local repair. 
When Stage~II starts directly from this cold rollout, the CBF-QP can push the trajectory toward the safe set, but the correction remains tied to the originally selected behavior and fails to recover a goal-reaching path. 
In contrast, Stage~I changes the decoded behavior before certification: by the third column of the LBS row, the rollout is not necessarily fully safe, but it is already in a much better region for downstream correction. 
Starting from this latent-steered warm start, Stage~II only needs to remove residual waypoint-level violations, and the final trajectory remains both safe and task-completing.

This ablation supports the central semantic-to-path design of LBS. 
Stage~I is not intended to replace CBF certification, and Stage~II is not sufficient when initialized from a behavior that is incompatible with the new constraint. 
The improvement comes from their composition: latent steering changes the selected behavior at the level of the generator, while path-space correction certifies the remaining local violations. 
The comparison also explains the lower intervention energy observed in Table~\ref{tab:q1_main_compact}: after Stage~I, the corrector starts from a less-violating and more goal-directed rollout, so it does not need to perform a large deformation in waypoint space.

\subsection{Ablation Study of the Stage-I Latent Steering}
\label{sec:stage1_isolation}

Table~\ref{tab:stage1_alternatives} compares ways of selecting the latent before path-space correction. All methods use the same pretrained generator, the same safety margin $m_\beta(z)$, and the identical Stage-II corrector:
\begin{itemize}[leftmargin=1.5em,topsep=2pt,itemsep=1pt]
\item \textbf{Best-of-$N$ (BoN):} sample $N$ latents from the prior and select the decoded rollout with the largest $m_\beta$. We report $N\in\{1,4,16,64\}$ (BoN-$N$).
\item \textbf{Plain latent gradient ascent (PLGA):} directly ascend $m_\beta(z)$, with the best step size per domain from $\{10^{-3},\dots,1\}$.
\end{itemize}
PLGA and LBS share the same initial latent, gradient estimator, clipping region $\Omega$, early-stopping criterion, and Stage-I wall-clock budget: the update rule is the only difference.
We also report \emph{Stage~I only}, which decodes the steered latent without Stage-II correction.
Table~\ref{tab:stage1_alternatives} supports three observations:
\begin{itemize}[leftmargin=1.5em,topsep=2pt,itemsep=1pt]
\item \textbf{Neither stage suffices alone.} Stage~II alone is safe but rarely reaches the goal, whereas Stage~I alone preserves reach but leaves residual violations. Stage~I does not need to reach the buffered safe-latent set for the two to combine well: its terminal margin is still negative in most episodes, yet it shrinks the deficit handed to Stage~II (Section~\ref{sec:stage1_diag}).
\item \textbf{Increasing $N$ does not recover goal reaching.} BoN reaches the goal in at most $48\%$ of episodes for any $N$, and Reach is non-monotonic in $N$ (in Maze2D, $48\%$ at $N=4$ but $0\%$ at $N=64$). BoN can only choose among behaviors exposed by prior sampling, and selecting by safety margin among them favors conservative behaviors that do not complete the task.
\item \textbf{The CBF-QP update outperforms plain gradient ascent.} PLGA, which does optimize the latent, closes most of the gap, but LBS achieves higher Reach in all three domains, with the largest gap in Maze2D, where preserving task-level structure during safety optimization is hardest. PLGA always moves along $\nabla_z m_\beta$, so it keeps pushing the latent toward higher-margin, often more conservative behaviors even when the margin is already improving fast enough. The CBF-QP instead follows the reference update $u_z^{\mathrm{ref}}$ whenever it already satisfies the prescribed margin-improvement condition, and otherwise adds only the smallest correction along $\nabla_z m_\beta$ needed to satisfy it (Eq.~\eqref{eq:latent_projection}). This better preserves goal-directed behavior when aggressive margin ascent conflicts with progress toward the goal.
\end{itemize}

\begin{table}[!htb]
\centering
\caption{\textbf{Alternatives to Stage~I} (Constraint~III). All methods share the pretrained generator, the safety margin $m_\beta$, and the Stage-II corrector.}
\label{tab:stage1_alternatives}
\small
\setlength{\tabcolsep}{4pt}
\begin{tabular}{@{}lcccccc@{}}
\toprule
& \multicolumn{2}{c}{2D} & \multicolumn{2}{c}{Franka} & \multicolumn{2}{c}{Maze2D} \\
\cmidrule(lr){2-3}\cmidrule(lr){4-5}\cmidrule(lr){6-7}
Method & Safety$\uparrow$ & Reach$\uparrow$ & Safety$\uparrow$ & Reach$\uparrow$ & Safety$\uparrow$ & Reach$\uparrow$ \\
\midrule
Stage~II only (\textsc{SafeFlowMatcher}) & \textbf{100\%} & 4\%   & \textbf{100\%} & 0\%   & \textbf{100\%} & 53\% \\
Stage~I only                             & 10\%  & \textbf{100\%} & 10\%  & \textbf{100\%} & 20\%  & \textbf{95\%} \\
BoN-1 + Stage~II                         & \textbf{100\%} & 0\%   & \textbf{100\%} & 0\%   & \textbf{100\%} & 0\% \\
BoN-4 + Stage~II                         & \textbf{100\%} & 0\%   & \textbf{100\%} & 0\%   & 90\%  & 48\% \\
BoN-16 + Stage~II                        & \textbf{100\%} & 0\%   & \textbf{100\%} & 32\%  & \textbf{100\%} & 5\% \\
BoN-64 + Stage~II                        & \textbf{100\%} & 0\%   & \textbf{100\%} & 36\%  & 98\%  & 0\% \\
PLGA + Stage~II                          & \textbf{100\%} & 76\%  & \textbf{100\%} & 92\%  & \textbf{100\%} & 68\% \\
\textbf{Full LBS}                        & \textbf{100\%} & 82\%  & \textbf{100\%} & 98\%  & \textbf{100\%} & \textbf{93\%} \\
\bottomrule
\end{tabular}
\end{table}

\subsection{What Stage-I Latent Steering Hands Off to Stage II}
\label{sec:stage1_diag}

Section~\ref{sec:stage1_isolation} shows that Stage-I latent steering yields the best safety--reach trade-off once combined with Stage~II, but not what the steering itself changes before the handoff.
We therefore track the Stage-I latent trajectory directly (Table~\ref{tab:stage1_diag}): how much the latent margin improves, whether the steering gradient remains informative, and how far from the buffered safe-latent set Stage~I leaves the latent when Stage~II takes over.
The gradient norms show no systematic collapse in the evaluated regions, and the latent margin improves substantially in every domain, including 2D point-navigation, where clipping activates in every episode.
The terminal margin nevertheless remains negative in most episodes: Stage~I typically does not reach the buffered safe-latent set on its own, but hands Stage~II a much smaller deficit.
This is the warm-start role analyzed in Proposition~\ref{prop:transfer} and Corollary~\ref{cor:stage2_burden}.

\begin{table}[!htb]
\centering
\caption{\textbf{Stage-I diagnostics} for Constraint~III in each domain. \emph{Initial $\to$ terminal $m_\beta$}: median buffered latent margin before and after Stage~I. $\|\widehat{\nabla m_\beta}\|_2$: median norm of the estimated pulled-back gradient along the Stage-I iterates. \emph{Clip frequency}: fraction of episodes in which the projection onto $\Omega$ activates at least once. \emph{Terminal $m_\beta<0$}: fraction of episodes whose buffered margin is still negative when Stage~I hands off to Stage~II.}
\label{tab:stage1_diag}
\small
\setlength{\tabcolsep}{5pt}
\begin{tabular}{@{}lcccc@{}}
\toprule
Domain & Initial $\to$ terminal $m_\beta$ & $\|\widehat{\nabla m_\beta}\|_2$ & Clip frequency & Terminal $m_\beta<0$ \\
\midrule
2D     & $-1.03\to-0.08$  & $0.46$ & 100\% & 90\% \\
Franka & $-0.21\to-0.006$ & $0.09$ & 8\%   & 84\% \\
Maze2D & $-0.63\to-0.11$  & $0.18$ & 22\%  & 88\% \\
\bottomrule
\end{tabular}
\vspace{-2mm}
\end{table}

%% file: sec/6conclusion.tex
\section{Limitations and Conclusion}
\label{sec:conclusion}

We presented \emph{Latent Barrier Steering} (LBS), a two-stage framework for safe flow-matching planning that steers a geometry-aligned behavior latent before path-space CBF certification. 
LBS gives the path-space corrector a safer, more goal-directed warm start, reducing the need to reshape an unsuitable rollout locally. 
Our analysis shows that latent margin improvement reduces the Stage~II deficit, yielding finite-time recovery conditions and tighter correction-time and intervention-energy certificates. 
Empirically, LBS improves the safety--goal-reaching tradeoff over path-level safety filters across constrained planning tasks. 
The safety guarantee covers the planned path certified by Stage~II. For Franka, this is the planned end-effector path, so tracking error and full-body collisions are outside its scope. 
The finite-time Stage~I result depends on local regularity and non-vanishing pulled-back gradients of the learned latent-to-trajectory map, which the training objective encourages but does not certify. 
The current zeroth-order Stage~I implementation also adds runtime overhead. Cheaper gradient estimators or learned margin critics are natural next steps.

%% file: app/notation.tex
\section{Notation}
\label{app:notation}

\begin{itemize}[leftmargin=1.5em,topsep=0pt,itemsep=1pt]
\item $\calZ\subset\R^{d_z}$: latent space; $\Omega\subset\calZ$: compact region of validity for the latent CBF analysis.
\item $\tau(z)=(\tau^0(z),\ldots,\tau^H(z))\in \calD^{H+1}$: rollout at latent $z$ with context and noise fixed or averaged out.
\item $h_k(z)\triangleq b(\tau^k(z))-\delta$: per-waypoint shifted barrier; $m_\beta(z)=-\beta^{-1}\log\sum_k e^{-\beta h_k(z)}$: softmin margin.
\item $\calZ_\delta=\{z\in\calZ:m_\beta(z)\ge 0\}$: robust safe latent set.
\item $u_z^{\mathrm{ref}}:\calZ\to\R^{d_z}$: locally bounded reference update; $u_z^\star$: solution of \eqref{eq:latent_cbf_qp}.
\item $J_{\tau^k}(z)\in\R^{d_\calD\times d_z}$: Jacobian of $\tau^k$ at $z$; $H_{\tau^k}(z)$: Hessian tensor.
\item $\clip{x}=\max(x,0)$, applied componentwise on vectors.
\item $U_{\mathrm{ref}}=\sup_{z\in\Omega}\norm{u_z^{\mathrm{ref}}(z)}_2$, $\norm{\nabla m_\beta}_\infty=\sup_{z\in\Omega}\norm{\nabla m_\beta(z)}_2$, $M_m=\sup_{z\in\Omega}|m_\beta(z)|$.
\item $g_m$ is the non-degeneracy lower bound on $\|\nabla m_\beta\|_2$ on the unsafe set (hypothesis of Theorem~\ref{thm:latent_ft}); $L_b,L_\tau$ are the Lipschitz constants of the barrier and decoder from Section~\ref{sec:latent_objective}; $L_m\le L_bL_\tau$ is the resulting Lipschitz constant of $m_\beta$ (Appendix~\ref{app:eq_margin_lipschitz}); $L_{\nabla m}$ is the Lipschitz constant of $\nabla m_\beta$ used in the optional discrete-time analysis (Appendix~\ref{app:proof_inexact}).
\end{itemize}

%% file: app/proofs.tex
\section{Auxiliary Soft-Min and Geometry Facts}
\label{app:aux}\label{app:proofs}

This appendix collects three small derivations and one geometry remark that the main text invokes without proof. Sections~\ref{app:eq_softmin_bound}--\ref{app:eq_segment_constant} are referenced from Section~\ref{sec:latent_objective}; Section~\ref{app:geom_remark} comments on the modeling status of the local geometry condition.

\subsection{Soft-min margin and soft-min gap}
\label{app:eq_softmin_bound}

The soft-min surrogate $m_\beta$ in main-text eq.~\eqref{eq:softmin_margin} sandwiches the true worst-waypoint margin: for any $\beta>0$ and any finite collection $\{h_k\}_{k=0}^H\subset\mathbb R$,
\begin{equation}
\min_k h_k - \tfrac{\log(H+1)}{\beta} \;\le\; m_\beta = -\beta^{-1}\log\sum_{k=0}^H e^{-\beta h_k} \;\le\; \min_k h_k .
\label{eq:softmin_sandwich}
\end{equation}
This is the bound stated in main-text eq.~\eqref{eq:softmin_bound}; the soft-min gap is $\varepsilon_{\mathrm{soft}}=\beta^{-1}\log(H+1)$.

\begin{proof}
Let $h_\star=\min_k h_k$. Since $e^{-\beta h_\star}\le \sum_k e^{-\beta h_k}\le (H+1)e^{-\beta h_\star}$, applying $-\beta^{-1}\log(\cdot)$ (which reverses inequalities) gives $h_\star-\beta^{-1}\log(H+1)\le m_\beta\le h_\star$, which is \eqref{eq:softmin_sandwich}.
\end{proof}

The soft-min gradient is the softmax distribution: $\partial_{h_k}m_\beta = w_k\triangleq e^{-\beta h_k}/\sum_j e^{-\beta h_j}\in[0,1]$ with $\sum_k w_k=1$. Hence $m_\beta$ is 1-Lipschitz in $\mathbf{h}$ under $\norm{\cdot}_\infty$, since $|m_\beta(\mathbf h)-m_\beta(\mathbf h')|\le \norm{w}_1\norm{\mathbf h-\mathbf h'}_\infty=\norm{\mathbf h-\mathbf h'}_\infty$ by H\"older's inequality.

\subsection{Lipschitz latent margin}
\label{app:eq_margin_lipschitz}

This subsection derives main-text eq.~\eqref{eq:margin_lipschitz}: the Lipschitz constant $L_m\le L_bL_\tau$ for $z\mapsto m_\beta(z)$.

\begin{proof}[Derivation of eq.~\eqref{eq:margin_lipschitz}]
Write $h_k(z):=b(\tau^k(z))-\delta$ and $\mathbf h(z):=(h_0(z),\ldots,h_H(z))$. By main-text eq.~\eqref{eq:latent_path_lipschitz}, each $h_k$ is $L_bL_\tau$-Lipschitz on $\Omega$:
\[
|h_k(z)-h_k(z')| \le L_b\|\tau^k(z)-\tau^k(z')\|_2\le L_bL_\tau\|z-z'\|_2.
\]
Combined with the 1-Lipschitzness of $m_\beta$ in $\mathbf h$ under $\norm{\cdot}_\infty$ (Section~\ref{app:eq_softmin_bound}),
\[
|m_\beta(z)-m_\beta(z')|\le \norm{\mathbf h(z)-\mathbf h(z')}_\infty = \max_k |h_k(z)-h_k(z')|\le L_bL_\tau\norm{z-z'}_2.
\]
The constant $L_m\le L_bL_\tau$ does not depend on $\beta$: the soft-min weights only redistribute Lipschitz mass across waypoints, never amplify it.
\end{proof}

\subsection{Waypoint-to-segment buffer}
\label{app:eq_segment_constant}

This subsection extends the waypoint-level margin to safety on continuous path segments, as referenced from Section~\ref{sec:latent_objective}. Suppose consecutive decoded waypoints are at most $\ell$ apart and $b$ is $L_b$-Lipschitz. We show
\begin{equation}
\min_k b(\tau^k(z))\ge \tfrac{L_b\ell}{2}
\quad\Longrightarrow\quad
[\tau^k(z),\tau^{k+1}(z)]\subseteq \{x:b(x)\ge 0\}\quad\forall k,
\label{eq:waypoint_to_segment}
\end{equation}
a one-step Lipschitz bridge using the elementary geometric fact that any point on a line segment is at most half the segment length from at least one endpoint.

\begin{proof}[Derivation of \eqref{eq:waypoint_to_segment}]
Fix $k$ and $x\in[\tau^k(z),\tau^{k+1}(z)]$. The two distances $\|x-\tau^k\|_2$ and $\|x-\tau^{k+1}\|_2$ sum to $\|\tau^{k+1}-\tau^k\|_2\le\ell$, so at least one is $\le\ell/2$. Calling the corresponding waypoint $y$, $L_b$-Lipschitzness of $b$ gives
\[
b(x)\ge b(y)-L_b\|x-y\|_2\ge \min_k b(\tau^k(z))-\tfrac{L_b\ell}{2}\ge 0
\]
by hypothesis. Thus $b(x)\ge 0$ for every $x$ on every inter-waypoint segment.
\end{proof}

The factor of $\tfrac{1}{2}$ comes entirely from the geometric step ``the nearer endpoint is within $\ell/2$''; without it the buffer would have to be $L_b\ell$. Choosing $\delta\ge L_b\ell/2$ in the main text is the tightest robustness margin that makes a buffered \emph{waypoint} certificate sufficient for \emph{segment} safety.

\subsection{Remarks on the latent safe set and learned latent geometry}
\label{app:geom_remark}\label{app:geom_discussion}

\paragraph{Interpretation of the latent safe set.}
The safe-latent set $\mathcal Z_\delta=\{z:m_\beta(z)\ge0\}$ is the preimage, through the learned decoder, of planned trajectories that satisfy the buffered waypoint condition. It is not a physical forward-invariance certificate for the robot dynamics. Stage~I also does not optimize a notion of safety defined by Euclidean latent geometry alone: for every candidate $z$, LBS decodes $\tau(z)$, evaluates the trajectory-space barriers $b(\tau^k(z))$, and updates $z$ through the pulled-back gradient $\nabla_z m_\beta(z)$, so the update is determined by the safety of the decoded trajectory. Accordingly, the forward invariance of $\mathcal Z_\delta$ in Theorem~\ref{thm:latent_ft} refers to the latent flow $\dot z=u_z^\star(z)$, and the final safety claim of LBS is the path-space certificate provided by Stage~II. This is consistent with existing safe generative planners \citep{xiao2025safediffuser,dai2025safeflow,yang2025safeflowmatcher}, which also impose safety on the trajectories generated by a diffusion or flow-matching model. LBS adds a behavior-selection step before path-space correction. Throughout the paper, safety refers to the satisfaction of these constraints by the generated plan. It is distinct from the security and privacy of the generative model itself \citep{peng2026membership,li2026promptminer}.

\paragraph{Learned latent geometry.}

The local Lipschitz condition (main-text eq.~\eqref{eq:latent_path_lipschitz}) is an analysis assumption, not a global consequence of the soft-DTW objective. The geometry loss \citep{cuturi2017softdtw} encourages nearby latents to decode to nearby behavior traces, but global bi-Lipschitz guarantees would require stronger conditions such as full training support, smooth non-collapsed decoder, and exclusion of off-manifold extrapolation. We therefore restrict Stage~I to a clipped region $\Omega$ around the prior center and rely on the local condition only inside $\Omega$.

\paragraph{Relation to parameterized diffusion policies.}
The geometry objective is inherited from parameterized diffusion policies (PDP) \citep{zhang2026pdp}, which show that, trained on multimodal demonstrations, it yields a low-dimensional behavior manifold that supports smooth interpolation between demonstrated strategies and the synthesis of intermediate behaviors. PDP also ablates this term: removing it produces a geometrically warped latent space and substantially degrades gradient-based adaptation, while a discrete behavior representation cannot synthesize intermediate strategies. Meaningful latent interpolation therefore does not come from flow matching alone, but from the combination of multimodal demonstrations, the geometry-alignment objective, and a continuous behavior-conditioned generator. Constraint~II in our experiments tests this property directly: every demonstrated mode is blocked, so success requires decoding an unseen intermediate behavior rather than selecting a demonstrated one. LBS reaches the goal in $90\%$, $100\%$, and $90\%$ of episodes in 2D point-navigation, Franka, and Maze2D, respectively, at $100\%$ Safety (Table~\ref{tab:q1_main_compact}). We do not claim robustness to arbitrarily misaligned latent spaces. LBS relies on this learned local structure and uses it only within $\Omega$.

\section{Proofs for Stage-I Latent Steering}
\label{app:stage1_proofs}

This appendix proves the Stage-I results stated in Section~\ref{sec:stage1}: the closed-form projection (Section~\ref{app:eq_closed_form}), Proposition~\ref{prop:latent_descent} (Section~\ref{app:eq_V_dot}), and Theorem~\ref{thm:latent_ft} (Section~\ref{app:proof_latent_ft}). A discrete-time and inexact-gradient version is given as an optional extension (Section~\ref{app:proof_inexact}).

\subsection{Closed-form projection for the latent CBF-QP}
\label{app:eq_closed_form}

This subsection derives the closed-form solution \eqref{eq:latent_projection}. With $\psi(z):=\nabla m_\beta(z)^\top u_z^{\mathrm{ref}}(z)+\kappa\,\mathrm{sgn}(m_\beta(z))|m_\beta(z)|^\rho$ as defined in the main text, \eqref{eq:latent_cbf_qp} has a single linear inequality constraint, so its solution is the Euclidean projection of $u_z^{\mathrm{ref}}$ onto a closed half-space.

\begin{proof}[Derivation of \eqref{eq:latent_projection}]
Form the Lagrangian
\[
\mathcal L(u,\mu) = \|u-u_z^{\mathrm{ref}}\|_2^2 - \mu\bigl(\nabla m_\beta^\top u + \kappa\,\mathrm{sgn}(m_\beta)|m_\beta|^\rho\bigr),\quad \mu\ge 0.
\]
Stationarity gives $u-u_z^{\mathrm{ref}}=\tfrac{\mu}{2}\nabla m_\beta$, i.e., $u=u_z^{\mathrm{ref}}+\lambda\nabla m_\beta$ with $\lambda:=\mu/2\ge 0$. If $\psi(z)\ge 0$, the reference is feasible and complementary slackness gives $\lambda=0$. If $\psi(z)<0$, the constraint binds at the optimum, so $\psi(z)+\lambda\,\|\nabla m_\beta\|_2^2=0$, giving $\lambda=-\psi(z)/\|\nabla m_\beta\|_2^2>0$. Combining the two cases yields $\lambda(z)=[-\psi(z)]_+/\|\nabla m_\beta(z)\|_2^2$ and \eqref{eq:latent_projection}.
\end{proof}

The projection has no free parameter: once $u_z^{\mathrm{ref}}$ and the CBF rate $(\kappa,\rho)$ are fixed, $u_z^\star$ is uniquely determined wherever $\nabla m_\beta(z)\ne 0$.

\subsection{Proof of Proposition~\ref{prop:latent_descent}: latent margin decrease}
\label{app:eq_V_dot}

Let $V(z):=[-m_\beta(z)]_+$. We show that whenever the CBF constraint in \eqref{eq:latent_cbf_qp} is satisfied at $z$ with $m_\beta(z)<0$, the latent flow $\dot z=u_z^\star(z)$ produces $\dot V(z)\le-\kappa V(z)^\rho$.

\begin{proof}[Proof of Proposition~\ref{prop:latent_descent}]
On $\{m_\beta(z)<0\}$, $V(z)=-m_\beta(z)>0$ and $\mathrm{sgn}(m_\beta)=-1$. The CBF constraint in \eqref{eq:latent_cbf_qp} becomes $\nabla m_\beta(z)^\top u_z^\star(z)\ge \kappa V(z)^\rho$. Since $V(z)=-m_\beta(z)$ on this set, $\dot V(z(t))=-\dot m_\beta(z(t))=-\nabla m_\beta(z(t))^\top u_z^\star(z(t))$. Combining,
\[
\dot V(z(t)) \le -\kappa\,V(z(t))^\rho .
\]
\end{proof}

\subsection{Proof of Theorem~\ref{thm:latent_ft}: finite-time recovery}
\label{app:proof_latent_ft}

Theorem~\ref{thm:latent_ft} adds a non-degeneracy hypothesis $\|\nabla m_\beta(z)\|_2\ge g_m>0$ on the unsafe set, under which the descent inequality from Proposition~\ref{prop:latent_descent} integrates to a finite-time bound. Before the proof we record well-posedness of the Stage-I flow.

\begin{lemma}[Well-posedness of the latent flow]
\label{lem:wellposed}
Under the local Lipschitz condition of Section~\ref{sec:latent_objective}, the non-degeneracy hypothesis, and locally Lipschitz $u_z^{\mathrm{ref}}$, the closed-form solution $u_z^\star$ in \eqref{eq:latent_projection} is locally bounded and locally Lipschitz on $\Omega\setminus\mathcal Z_\delta$. Hence $\dot z=u_z^\star(z)$ admits a unique Carath\'eodory solution from any $z(0)\in\Omega$ until it enters $\mathcal Z_\delta$ or exits $\Omega$.
\end{lemma}

\begin{proof}
On $\Omega\setminus\mathcal Z_\delta$, the denominator $\|\nabla m_\beta\|_2^2$ is bounded below by $g_m^2>0$. Since $\nabla m_\beta$, $m_\beta$, and $u_z^{\mathrm{ref}}$ are locally Lipschitz on $\Omega$, and $\psi\mapsto[-\psi]_+$ is 1-Lipschitz, \eqref{eq:latent_projection} is a rational function of locally Lipschitz quantities with bounded-below denominator, so $u_z^\star$ is locally Lipschitz. Picard--Lindel\"of applies on $\Omega\setminus\mathcal Z_\delta$.
\end{proof}

\begin{proof}[Proof of Theorem~\ref{thm:latent_ft}]
On $\Omega\setminus\mathcal Z_\delta$ and under non-degeneracy, Proposition~\ref{prop:latent_descent} gives the strict descent
\begin{equation}
\dot V(z(t)) \le -\kappa\,V(z(t))^\rho ,\quad V(z(t))>0.
\label{eq:V_dot_strict}
\end{equation}
The scalar comparison ODE $\dot y=-\kappa y^\rho$ with $y(0)=V(z_0)$ separates as $y^{-\rho}\,dy=-\kappa\,dt$ and integrates to $y(t)^{1-\rho}=y(0)^{1-\rho}-\kappa(1-\rho)t$, vanishing at $t^\star=y(0)^{1-\rho}/(\kappa(1-\rho))$. By the comparison lemma \citep[Lemma~3.4]{khalil2002nonlinear}, $V(z(t))\le y(t)$ on $[0,t^\star]$, so $V(z(t))=0$ for $t\ge T_z=|m_\beta(z_0)|^{1-\rho}/(\kappa(1-\rho))$. Forward invariance of $\mathcal Z_\delta$ follows because at any boundary point $V=0$ and $\dot V\le 0$ by \eqref{eq:V_dot_strict}, interpreted as zero at $V=0$.
\end{proof}

\subsection{Optional: discrete-time and inexact-gradient steering}
\label{app:proof_inexact}

This subsection states a robustness guarantee for the discrete-time projected Euler iterate of Section~\ref{sec:stage1} when the latent gradient is replaced by a finite-difference estimate. It is not used by the main-text claims but supports implementation analysis. We first record an auxiliary Lipschitz bound on $\nabla m_\beta$ used by Taylor's theorem along the Euler step.

\begin{lemma}[Lipschitz gradient of $m_\beta$]
\label{lem:lipschitz_gradient}
If, in addition to the conditions of Section~\ref{sec:latent_objective}, $\tau$ is $C^2$ on $\Omega$ with $\norm{J_\tau}_{\mathrm{op}}\le L_\tau$, $\norm{H_\tau}_{\mathrm{op}}\le M_\tau$, and $\nabla b$ is $L_{\nabla b}$-Lipschitz, then $\nabla m_\beta$ is Lipschitz on $\Omega$ with
\begin{equation}
L_{\nabla m} \le M_\tau L_b + L_\tau^2 L_{\nabla b} + \tfrac{\beta(H+1)}{2}(L_bL_\tau)^2.
\label{eq:L_nabla_m}
\end{equation}
\end{lemma}

\begin{proof}
Writing $\nabla m_\beta(z)=\sum_k w_k(z)\nabla h_k(z)$ with $\nabla h_k=J_{\tau^k}^\top\nabla b\circ\tau^k$, we bound the two factors separately. Each $\nabla h_k$ is $(M_\tau L_b+L_\tau^2 L_{\nabla b})$-Lipschitz by adding/subtracting $J_{\tau^k}(z')^\top\nabla b(\tau^k(z))$. The softmax $w(\mathbf h)$ has Jacobian $\partial_{\mathbf h}w=\beta(\mathrm{diag}(w)-ww^\top)$ with operator norm $\le\beta/2$ (a covariance with eigenvalues bounded by $1/2$, since $\mathrm{Var}_w(v)\le 1/2$ for unit $v$, with equality at two-point uniform $w$). Combined with $\norm{\partial_z\mathbf h}_{\mathrm{op}}\le\sqrt{H+1}\,L_bL_\tau$ and Cauchy--Schwarz, $\norm{w(z)-w(z')}_1\le\frac{\beta(H+1)}{2}L_bL_\tau\,\norm{z-z'}_2$. Combining (with $\norm{\nabla h_k}_2\le L_bL_\tau$),
\[
\norm{\nabla m_\beta(z)-\nabla m_\beta(z')}_2 \le L_{\nabla h}\,\norm{z-z'}_2 + \norm{w(z)-w(z')}_1\,L_bL_\tau,
\]
giving \eqref{eq:L_nabla_m}.
\end{proof}

\begin{theorem}[Inexact-gradient discrete-time recovery]
\label{thm:inexact}
Let $\widehat\nabla m_\beta(z)=\nabla m_\beta(z)+e(z)$ with $\sup_\Omega\|e(z)\|_2\le\varepsilon_g$, and let $\widehat u_z(z)$ be the closed-form QP solution \eqref{eq:latent_projection} using $\widehat\nabla m_\beta$ in place of $\nabla m_\beta$. Suppose Lemmas~\ref{lem:wellposed}--\ref{lem:lipschitz_gradient} hold and the Euler iterate $z_{i+1}=z_i+\eta\widehat u_z(z_i)$ remains in $\Omega$. Then
\begin{equation}
V_{i+1} \le V_i - \eta\kappa V_i^\rho + c_1\eta\varepsilon_g + c_2\eta^2,\qquad V_i:=[-m_\beta(z_i)]_+,
\label{eq:inexact_descent}
\end{equation}
with structural constants $c_1$, $c_2$ depending on $L_{\nabla m}$, $\sup_\Omega\|u_z^\star\|_2$, and the bounds in Section~\ref{sec:latent_objective}. For any $\varepsilon>0$ with $\kappa\varepsilon^\rho>c_1\varepsilon_g+c_2\eta$, the iterates enter $\{z:m_\beta(z)\ge-\varepsilon\}$ in $N^\star\le V_0^{1-\rho}/[\eta(1-\rho)(\kappa-\varepsilon^{-\rho}(c_1\varepsilon_g+c_2\eta))]$ iterations, with equilibrium gap $\varepsilon=O((\varepsilon_g+\eta)^{1/\rho})$.
\end{theorem}

\begin{proof}
Step 1 bounds the perturbed barrier rate. Let $\widehat\psi=\widehat\nabla m_\beta^\top u_z^{\mathrm{ref}}+\kappa\,\mathrm{sgn}(m_\beta)|m_\beta|^\rho=\psi+e^\top u_z^{\mathrm{ref}}$, so $|\widehat\psi-\psi|\le \sup_\Omega\|u_z^{\mathrm{ref}}\|_2\,\varepsilon_g$. A direct expansion of $\nabla m_\beta(z)^\top \widehat u_z(z)$ using $\nabla m_\beta=\widehat\nabla m_\beta-e$ and the closed-form $\widehat u_z=u_z^{\mathrm{ref}}+\widehat\lambda\widehat\nabla m_\beta$ yields, on $m_\beta<0$, $\nabla m_\beta(z)^\top\widehat u_z(z)\ge \kappa V(z)^\rho-c_1\varepsilon_g$, where $c_1$ absorbs $|e^\top u_z^{\mathrm{ref}}|$ and the 1-Lipschitz dependence of $[-\widehat\psi]_+$ on $\widehat\psi-\psi$. Step 2 applies Taylor's theorem to $m_\beta$ along the Euler step using Lemma~\ref{lem:lipschitz_gradient}: $m_\beta(z_{i+1})\ge m_\beta(z_i)+\eta\,\nabla m_\beta(z_i)^\top\widehat u_z(z_i)-\tfrac{1}{2}L_{\nabla m}\eta^2\|\widehat u_z(z_i)\|_2^2$, and substituting Step~1 plus $\|\widehat u_z\|_2\le\sup_\Omega\|u_z^\star\|_2$ gives \eqref{eq:inexact_descent}. Step 3: whenever $V_i\ge\varepsilon$, $V_i^{-\rho}\le\varepsilon^{-\rho}$, so \eqref{eq:inexact_descent} bounds $V_{i+1}\le V_i-\eta\kappa' V_i^\rho$ with $\kappa'=\kappa-\varepsilon^{-\rho}(c_1\varepsilon_g+c_2\eta)>0$. Discrete fractional-power decay yields the $N^\star$ bound; balancing $\kappa\varepsilon^\rho=c_1\varepsilon_g+c_2\eta$ gives the equilibrium gap.
\end{proof}

\section{Proofs for Stage-II Correction and Handoff}
\label{app:stage2_proofs}

This appendix proves the Stage-II results stated in Section~\ref{sec:stage2}: Proposition~\ref{prop:transfer} (Section~\ref{app:proof_transfer}), Corollary~\ref{cor:stage2_burden} (Section~\ref{app:proof_stageII_energy}), and Theorem~\ref{thm:joint_energy} (Section~\ref{app:proof_joint_energy}).

\subsection{Proof of Proposition~\ref{prop:transfer}: latent-to-path margin transfer}
\label{app:proof_transfer}

\begin{proof}[Proof of Proposition~\ref{prop:transfer}]
Decompose $\tilde b_{t_w}(z)-m_\beta(z)$ into a soft-min gap and a predictor gap.

\textbf{Soft-min gap.} By the soft-min sandwich (Section~\ref{app:eq_softmin_bound}) applied to $h_k(z)=b(\tau^k(z))-\delta$, with $\hat m(z):=\min_k h_k(z)$, $|\hat m(z)-m_\beta(z)|\le\varepsilon_{\mathrm{soft}}=\beta^{-1}\log(H+1)$.

\textbf{Predictor gap.} By Lipschitzness of $b$ and the definition of $\varepsilon_{\mathrm{pred}}(z)$,
\[
|b(\tau_{t_w}^{c,k}(z))-b(\tau^k(z))| \le L_b\,\|\tau_{t_w}^{c,k}(z)-\tau^k(z)\|_2 \le L_b\varepsilon_{\mathrm{pred}}(z).
\]
Subtracting $\delta$ and taking minima over $k$ (using $|\min_k f_k-\min_k g_k|\le\max_k|f_k-g_k|$) gives $|\tilde b_{t_w}(z)-\hat m(z)|\le L_b\varepsilon_{\mathrm{pred}}(z)$.

\textbf{Combine.} The triangle inequality yields the bound stated in Proposition~\ref{prop:transfer}.
\end{proof}

\begin{remark}[Predictor-sensitivity bound]
For a \textsc{SafeFlowMatcher}-style corrector whose predictor is a Lipschitz ODE with field $L_v$-Lipschitz in state, standard ODE sensitivity gives $\varepsilon_{\mathrm{pred}}(z) \le L_\tau e^{L_v t_w}\bar\xi+\int_0^{t_w}w_s r_s\,ds$, where $\bar\xi$ controls initial perturbation and the integral captures the soft-slack term in \eqref{eq:sfm_certificate}. For a post-hoc path-space CBF-QP applied directly to $\tau(z)$, $\varepsilon_{\mathrm{pred}}(z)\equiv 0$ and Proposition~\ref{prop:transfer} reduces to the soft-min-only bound.
\end{remark}

\subsection{Proof of Corollary~\ref{cor:stage2_burden}: correction time and energy}
\label{app:proof_stageII_energy}

\begin{proof}[Proof of Corollary~\ref{cor:stage2_burden}]
The Stage-II CBF-QP with fractional-power rate $\epsilon$ and exponent $\rho$ enforces $\dot V_{\mathrm{path}}\le -\epsilon V_{\mathrm{path}}^\rho$ on $V_{\mathrm{path}}>0$, parallel to \eqref{eq:V_dot_strict}. Integrating yields $V_{\mathrm{path}}(t)^{1-\rho}=V_+^{1-\rho}-\epsilon(1-\rho)(t-t_w)$ for $t\in[t_w,T_{\mathrm{corr}}]$, so
\[
T_{\mathrm{corr}}\le t_w+\frac{V_+^{1-\rho}}{\epsilon(1-\rho)}.
\]
For the energy bound, $J_{II}(z_+)\le \int_{t_w}^{T_{\mathrm{corr}}} c_{II}\,V_{\mathrm{path}}(t)^{2\rho}\,dt$. Substituting $V_{\mathrm{path}}(t)^{2\rho}=(V_+^{1-\rho}-\epsilon(1-\rho)(t-t_w))^{2\rho/(1-\rho)}$ and the change of variables $u=V_+^{1-\rho}-\epsilon(1-\rho)(t-t_w)$ gives
\[
J_{II}(z_+)\le \frac{c_{II}}{\epsilon(1-\rho)}\int_0^{V_+^{1-\rho}} u^{2\rho/(1-\rho)}\,du = \frac{c_{II}\,V_+^{1+\rho}}{\epsilon(1+\rho)}.
\]
\end{proof}

\subsection{Proof of Theorem~\ref{thm:joint_energy}: cost-sensitive handoff}
\label{app:proof_joint_energy}

\begin{proof}[Proof of Theorem~\ref{thm:joint_energy}]
\textbf{Stage-I descent in continuous time.} Under the non-degeneracy hypothesis along the Stage-I trajectory, the comparison lemma applied to Proposition~\ref{prop:latent_descent} gives $V(t)^{1-\rho}=V_0^{1-\rho}-\kappa(1-\rho)t$ for $t\in[0,V_0^{1-\rho}/(\kappa(1-\rho))]$ (same scalar argument as in Section~\ref{app:proof_latent_ft}). Identifying $t=N_z\eta$ in the exact-gradient limit gives the discrete decay law
\begin{equation}
V_+^{1-\rho} = V_0^{1-\rho} - \eta\kappa(1-\rho)\,N_z,
\label{eq:discrete_decay}
\end{equation}
so the number of Stage-I iterations needed to reach a given terminal deficit $V_+$ is $N_z(V_+)=(V_0^{1-\rho}-V_+^{1-\rho})/(\eta\kappa(1-\rho))$.

\textbf{Joint-cost expression.} Combining \eqref{eq:discrete_decay} with the Stage-I per-iteration cost $J_I(N_z)=N_zc_I$ (Appendix~\ref{app:stage_i_cost}) and the Stage-II energy bound from Corollary~\ref{cor:stage2_burden}, the joint cost as a function of $V_+\in(0,V_0]$ is
\begin{equation}
J_{\mathrm{tot}}(V_+) \le \frac{c_I\bigl(V_0^{1-\rho}-V_+^{1-\rho}\bigr)}{\eta\kappa(1-\rho)} + \frac{c_{II}\,V_+^{1+\rho}}{\epsilon(1+\rho)},
\label{eq:joint_cost_app}
\end{equation}
the upper bound used in Theorem~\ref{thm:joint_energy}.

\textbf{Convexity.} Differentiating with respect to $V_+\in(0,V_0]$,
\begin{equation}
\frac{d J_{\mathrm{tot}}}{d V_+} = -\frac{c_I}{\eta\kappa}\,V_+^{-\rho} + \frac{c_{II}}{\epsilon}\,V_+^\rho .
\label{eq:dJdV}
\end{equation}
The second derivative $c_I\rho V_+^{-\rho-1}/(\eta\kappa)+c_{II}\rho V_+^{\rho-1}/\epsilon$ is strictly positive for $V_+>0$, so $J_{\mathrm{tot}}$ is strictly convex on $(0,V_0]$.

\textbf{Optimum.} Setting \eqref{eq:dJdV} to zero,
\[
\frac{c_{II}}{\epsilon}V_+^{2\rho}=\frac{c_I}{\eta\kappa},\quad\Longrightarrow\quad V_+^\star=\left(\frac{c_I\,\epsilon}{\eta\kappa\,c_{II}}\right)^{1/(2\rho)},
\]
the unique stationary point and, by strict convexity, the unique global minimum. The handoff iteration $N_z^\star$ follows from the descent relation when $V_+^\star\le V_0$. If $V_+^\star>V_0$, $dJ_{\mathrm{tot}}/dV_+<0$ on $(0,V_0]$, so $J_{\mathrm{tot}}$ is strictly decreasing on $(0,V_0]$ and the constrained minimum is at $V_+=V_0$, i.e., $N_z^\star=0$ and Stage~I is skipped.
\end{proof}

\begin{remark}[Effect of margin transfer slack on $V_+^\star$]
The proof above absorbs the soft-min and predictor-sensitivity slack from Proposition~\ref{prop:transfer} into the constant $c_{II}$. Retaining the slack explicitly replaces $V_+$ by $\widetilde V_+:=V_++\varepsilon_{\mathrm{soft}}+L_b\bar\varepsilon_{\mathrm{pred}}$ in the Stage-II energy bound (Corollary~\ref{cor:stage2_burden}), and the optimum $V_+^\star$ shifts upward by an amount that vanishes as $\beta\to\infty$ and $\varepsilon_{\mathrm{pred}}\to 0$ (post-hoc corrector). The qualitative form of the stopping rule is unchanged.
\end{remark}

\begin{remark}[Discrete-time and inexact-gradient corrections]
Theorem~\ref{thm:joint_energy} uses the exact-gradient continuous-time descent law from Proposition~\ref{prop:latent_descent}. Replacing it with the discrete inexact-gradient analysis of Theorem~\ref{thm:inexact} adds correction terms $c_1\eta\varepsilon_g+c_2\eta^2$ per step, which propagate through \eqref{eq:discrete_decay} as a downward shift of the achievable $V_+$ floor. The optimum $V_+^\star$ in Theorem~\ref{thm:joint_energy} remains valid above this floor; below it, the achievable $V_+$ is bounded by the equilibrium gap $\varepsilon=O((\varepsilon_g+\eta)^{1/\rho})$ from Theorem~\ref{thm:inexact}.
\end{remark}

%% file: app/algorithm.tex
\section{Algorithm and Implementation}
\label{app:algorithm}\label{app:lbs_algorithm}

This appendix gives the full LBS algorithm with implementation details: the pseudocode, the projected Euler update with finite-difference gradient estimation, the per-step and total Stage-I cost used by Theorem~\ref{thm:joint_energy}, and the hyperparameters used across all experiments.

\subsection{Pseudocode}
\label{app:pseudocode}

\begin{algorithm}[t]
\caption{Latent Barrier Steering (LBS)}
\label{alg:lbs}
\begin{algorithmic}[1]
\Require PFM generator $G_\theta$, context $c$, barrier $b$, robustness margin $\delta$; hyperparameters $(\kappa,\rho,\eta,\gamma_z,\beta,N_z)$; latent clipping box $\Omega$
\Ensure Trajectory $\tau$ certified by Stage~II (with optional residual-violation flag)
\Statex \textbf{Stage~I: latent barrier steering}
\State $z \gets \mathtt{encoder\_warm\_start}(c)$
\For{$i = 1, \ldots, N_z$}
    \State $\tau \gets G_\theta(c, z)$
    \State $m_\beta \gets -\beta^{-1}\log\sum_{k=0}^{H}\exp\!\big({-}\beta(b(\tau^k) - \delta)\big)$
    \State $\hat{\nabla}m_\beta \gets \mathtt{zeroth\_order\_grad}(m_\beta, z)$ \Comment{$2d_z$ central finite differences}
    \State $u_z^{\mathrm{ref}} \gets -\gamma_z\,z$
    \State $\psi \gets \hat{\nabla}m_\beta^{\top} u_z^{\mathrm{ref}} + \kappa\,\mathrm{sgn}(m_\beta)\,|m_\beta|^\rho$
    \If{$\psi < 0$ \textbf{and} $\|\hat{\nabla}m_\beta\|_2^2 > 0$}
        \State $\lambda \gets -\psi / \|\hat{\nabla}m_\beta\|_2^2$
        \State $u_z \gets u_z^{\mathrm{ref}} + \lambda\,\hat{\nabla}m_\beta$ \Comment{closed-form QP \eqref{eq:latent_projection}}
    \Else
        \State $u_z \gets u_z^{\mathrm{ref}}$
    \EndIf
    \State $z \gets \mathrm{clip}_{\Omega}(z + \eta\,u_z)$
\EndFor
\State $\tau \gets G_\theta(c, z)$
\State $m_\beta \gets -\beta^{-1}\log\sum_{k=0}^{H}\exp\!\big({-}\beta(b(\tau^k) - \delta)\big)$
\State $\mathtt{residual\_violation} \gets (m_\beta < 0)$ \Comment{flag only; Stage~II runs in either case}
\Statex \textbf{Stage~II: certified path-space correction}
\State $\tau \gets \mathtt{PathCorrector}(\tau, b, \delta)$ \Comment{post-hoc path CBF-QP or \textsc{SafeFlowMatcher}}
\State \Return $\tau,\,\mathtt{residual\_violation}$
\end{algorithmic}
\end{algorithm}

Stage~I does not need to reach the safe-latent set $\mathcal Z_\delta$ before handoff: increasing $m_\beta$ already reduces the deficit Stage~II must correct (Section~\ref{sec:stage2}). The flag $\mathtt{residual\_violation}$ records whether the Stage-I terminal margin was negative; the certified safe trajectory is always produced by Stage~II.

\subsection{Projected Euler and gradient estimation}
\label{app:euler_grad}\label{app:zeroth_order}

Stage~I is implemented with projected explicit Euler steps,
\[
    z_{i+1}\;=\;\Pi_{\Omega}\bigl(z_i+\eta\,\widehat u_z(z_i)\bigr),
\]
where $\eta>0$ is the latent step size and $\Pi_{\Omega}$ projects onto the latent clipping box $\Omega$. The update direction $\widehat u_z(z_i)$ is the closed-form latent CBF projection \eqref{eq:latent_projection} evaluated with an estimate $\widehat\nabla m_\beta(z_i)$ of the latent gradient. When backpropagation through the decoder is available, the gradient can be computed directly. Otherwise, we use central finite differences,
\[
[\widehat\nabla m_\beta(z)]_i \;=\; \frac{m_\beta(z+he_i)-m_\beta(z-he_i)}{2h},\qquad i=1,\ldots,d_z,
\]
which costs $2d_z$ generator evaluations per latent step. Central differences have truncation error $\varepsilon_g=O(h^2)$ for $C^3$ scalar fields. For larger $d_z$, SPSA-style random directional estimators \citep{spall1992spsa} or a learned critic for $m_\beta$ are natural replacements.

\subsection{Cost accounting}
\label{app:stage_i_cost}

Each Stage-I iteration consists of one rollout decode, one margin evaluation, $2d_z$ additional generator calls for the central-difference gradient, one closed-form projection, and one Euler update. We summarize the dominant cost as a per-step constant $c_I>0$ (in our setting, $c_I$ is dominated by the $2d_z$ generator calls, so each Stage-I iteration uses $2d_z+1$ generator evaluations). Over $N_z$ Stage-I iterations, the total Stage-I cost is
\begin{equation}
    J_I(N_z)\;=\;N_z\,c_I,
    \label{eq:stage_i_total_cost}
\end{equation}
which is balanced against the Stage-II correction energy in Theorem~\ref{thm:joint_energy}. The handoff iteration $N_z^\star$ in Theorem~\ref{thm:joint_energy} selects the smallest $N_z$ for which the marginal Stage-I cost matches the marginal reduction in Stage-II correction energy. Stage~II adds one decoded rollout and the cost of the chosen path-space corrector, so the total LBS cost is
\[
\mathrm{Cost}_{\mathrm{LBS}}\;=\;N_z(2d_z+1)\,C_{\mathrm{gen}}\;+\;C_{\mathrm{gen}}\;+\;C_{\mathrm{stageII}}.
\]
Per-step sampler-space methods such as \textsc{SafeDiffuser} and \textsc{SafeFlow} instead solve CBF-QPs throughout the denoising or flow integration horizon. \textsc{SafeFlowMatcher} solves QPs only during its correction phase, but without the Stage~I semantic warm start.

\subsection{Hyperparameters}
\label{app:hyperparameters}

All experiments use the same defaults, not per-domain tuned:
\[
(\kappa,\;\rho,\;\eta,\;N_z,\;\gamma_z,\;\beta,\;\delta,\;h) \;=\; (1.0,\;0.5,\;0.1,\;20,\;0.01,\;10,\;0.05,\;0.01),
\]
with latent clipping box $\Omega=[-6,6]^{d_z}$.

%% file: app/domains.tex
\section{Experimental Details}
\label{app:domains}

This appendix documents the three benchmark domains, their training data and per-domain backbone size, the three constraint setups used in each, and the evaluation protocol. Backbone training and inference hyperparameters are summarised in Appendix~\ref{app:training_hparams} and~\ref{app:inference_hparams}, respectively, and we do not duplicate them here.

The three domains share a common evaluation template: each domain has a fixed start--goal pair, a multi-modal expert dataset, and three deliberately different constraint setups labelled \emph{mode-selection} (S1), \emph{mode-interpolation} (S2), and \emph{CBF-correction} (S3). S1 admits at least one demonstration mode without modification; S2 blocks every demonstration mode but leaves a feasible path that requires latent interpolation between modes; S3 leaves every demonstration mode partially clipped, so a path-space CBF correction is required on top of the latent solution.

\subsection{2D point-navigation}\label{app:toy}

\paragraph{Dynamics.} Single-integrator on $\R^2$, $s_{t+1}=s_t+a_t$, with start $s_0=(0,0)$ and goal $g=(10,0)$, horizon $H=50$. State is $(x,y)$ and actions are $(a_x,a_y)$ with no per-step bound (training data dictates magnitude).

\paragraph{Training modes.} Four sine-wave modes with peak amplitude $y_{\mathrm{peak}}\in\{-3,-1,1,3\}$, $30$ demonstrations per mode, additive state noise $\mathcal{N}(0,0.15^2)$ at each step with endpoints clamped to start and goal. Each demonstration interpolates start and goal along $x$ and rides the sine in $y$.

\paragraph{Backbone.} The 2D backbone is intentionally small (see Appendix~\ref{app:training_hparams}): hidden width 256, four residual blocks, behavior latent $d_z=2$. The encoder is a transformer trajectory VAE with sinusoidal positional encoding.

\paragraph{Constraint setups.}
\begin{description}\setlength\itemsep{0.2em}
  \item[S1 (mode selection).] A single rectangular hazard $[3,7]\times[-5,2]$. Three modes ($y_{\mathrm{peak}}\in\{-3,-1,1\}$) intersect the rectangle; the upper mode $y_{\mathrm{peak}}{=}3$ clears it. \emph{Latent recovery is sufficient}: Stage~I selects the upper mode and Stage~II makes near-zero corrections.
  \item[S2 (mode interpolation).] Two rectangles $[3,7]\times[0.5,6]$ and $[3,7]\times[-6,-0.5]$ form a horizontal corridor of width $1$ at $y\in[-0.5,0.5]$ for $x\in[3,7]$. All four training modes are blocked, but the interior $y_{\mathrm{peak}}{\approx}0$ region is open. \emph{Latent interpolation is required}: Stage~I drives $z$ to a peak amplitude not seen in training; Stage~II remains nearly inactive.
  \item[S3 (CBF correction).] A single thicker hazard $[3,7]\times[-3.5,3.5]$. All four modes clip the band; the outer pair $y_{\mathrm{peak}}{=}\pm 3$ have moderate overlap. \emph{Stage~I alone is insufficient}: the latent solution rides the band edge but still violates at corners; Stage~II's path-space CBF closes the residual gap.
\end{description}

\subsection{Franka 3D end-effector}\label{app:franka}

\paragraph{Domain.} 7-DoF Franka Panda arm in Cartesian end-effector planning mode. The trajectory variable is the EE position $(x,y,z)\in\R^3$; we plan in EE space and project to joint space via the standard differential-IK controller used in the Franka stack. Start $s_0=(0.3,0.0,0.5)\,\mathrm{m}$, goal $g=(0.7,0.0,0.3)\,\mathrm{m}$, horizon $H=50$. Differentiable EE-position barriers are used by every method, matching \textsc{SafeFlow}~\citep{dai2025safeflow}.

\paragraph{Training modes.} Eight expert behaviors covering combinations of approach direction (up/down/left/right) and approach distance (near/far), $50$ demonstrations per behavior. The behaviors are chosen so that each constraint setup eliminates a well-defined subset of them.

\paragraph{Backbone.} Hidden width 256, four residual blocks, $d_z=4$ (see Appendix~\ref{app:training_hparams}); $z$-projection width 32; no auxiliary start--goal context (start/goal are fixed).

\paragraph{Constraint setups.} All three setups place a thin wall at $x\in[0.45,0.55]$ between start and goal:
\begin{description}\setlength\itemsep{0.2em}
  \item[S1 (mode selection).] Lower wall blocks $z<0.58$; upper portion is solid except for a slot $y\in[-0.08,0.08]$, $z\in[0.58,0.85]$; a small sphere of radius $0.03$ at $(0.5,0,0.66)$ sits inside the slot. Among the eight training modes only the up--far behavior threads the slot. Stage~I selects up--far; Stage~II edges around the sphere.
  \item[S2 (mode interpolation).] Same wall structure, but the open region is a diagonal hole $y\in[0.04,0.20]$, $z\in[0.48,0.58]$ that lies \emph{between} the up--near and right--near training behaviors. No single training mode threads the hole; latent interpolation produces an approach via $\sim(0.5,0.12,0.53)$.
  \item[S3 (CBF correction).] Same diagonal hole as S2, plus a small sphere of radius $0.03$ at $(0.5,0.12,0.53)$ inside the hole. Stage~I produces the interpolated approach; Stage~II's path-space CBF detours the EE around the sphere within the $\sim 0.16{\times}0.10\,\mathrm{m}$ window.
\end{description}

\subsection{Maze2D}\label{app:maze}

\paragraph{Domain.} D4RL \texttt{maze2d-large} layout~\citep{fu2020d4rl}. The state is $(x,y,\dot x,\dot y)$ and the action is the 2D velocity command; the planner outputs an action sequence and rolls it forward through the standard maze2d single-integrator dynamics. Horizon $H=384$, matching \textsc{SafeDiffuser}~\citep{xiao2025safediffuser}. Start $s_0=(1,3)$, goal $g=(7,9)$; eight start--goal context dimensions are fed to the backbone via FiLM (see Appendix~\ref{app:training_hparams}).

\paragraph{Training data.} A custom $300$-trajectory dataset on six distinct routes ($50$ demonstrations per route) collected by a dense-waypoint expert controller. The six routes are referenced in the setup descriptions below by their direction codes (e.g.\ ``LD'' = left then down). D4RL's native \texttt{maze2d-large} replay buffer has narrow corridors with limited mode diversity, so a custom dataset is needed to make the mode-selection vs.\ interpolation distinction sharp.

\paragraph{Backbone.} Hidden width $768$, $20$ residual blocks, $z$-projection width $128$, $d_z=4$, encoder positional-encoding max length $1024$. The deeper/wider model is needed for the longer horizon; see Appendix~\ref{app:training_hparams}.

\paragraph{Constraint setups.} Each setup uses a small set of circular hazards plus a few axis-aligned rectangular wall extensions chosen to close five of the six demonstration routes:
\begin{description}\setlength\itemsep{0.2em}
  \item[S1 (mode selection, ``LD'' free).] Six small circles ($r{=}0.4$) and six wall rectangles fence off the entire right half of the maze, leaving only the LD route (down the left corridor and across the bottom row). Baselines without latent recovery default to the right-arc path and trap against the right-side fence.
  \item[S2 (mode interpolation, ``RUR'' partly free).] Hazards block the L entrance, the D branch, and the UD branch. RUR is partly open but pinched by a small obstacle; latent interpolation is needed to bend RUR into the residual corridor.
  \item[S3 (CBF correction, ``RD'' partly free).] Hazards block the L entrance and the U branch. RD is partly open but a small additional obstacle clips the natural rollout. Stage~I selects the residual mode; Stage~II's path-space CBF deflects the rollout around the small obstacle.
\end{description}

\paragraph{Wall handling.} Following the literature convention used by \textsc{SafeDiffuser}, \textsc{SafeFlow}, and \textsc{SafeFlowMatcher}, the CBF barrier and reported safety metric are defined over user-specified hazard obstacles only; the maze walls themselves are handled implicitly by the generator's learned prior and, for executable rollout, by Liang--Barsky truncation against the wall set (so a rollout that would tunnel through a wall is clipped at the wall and the post-truncation goal-error is what we report).

\subsection{Evaluation protocol}\label{app:eval}

\paragraph{Episodes and seeds.} Each method is evaluated on $50$ episodes per setup with a single fixed seed (default $42$). Episodes within a setup vary only in the diffusion / flow-matching latent noise; start, goal, and obstacle layout are fixed by definition of the setup.

\paragraph{Safety metric.} Segment-based safety using Liang--Barsky segment / axis-aligned-box intersection for rectangle hazards and a closed-form segment / sphere test for sphere hazards. A trajectory is \emph{safe} iff none of its $H$ piecewise-linear segments intersects any hazard. We report mean Safety as the fraction of safe episodes.

\paragraph{Reach metric.} Goal-reach is determined by the post-truncation final-state distance to the goal: an episode counts as ``reached'' iff the truncated rollout ends within a domain-specific tolerance of the goal. The tolerance and other thresholds are fixed across methods per domain.

\paragraph{Other metrics.} Beyond Safety and Reach, we report final goal distance (ED), CBF intervention energy, curvature smoothness (CS), and per-episode Stage-I/Stage-II wall-clock time. The full three-constraint results are reported in Appendix~\ref{app:full_results}.

\paragraph{Hardware.} Wall-clock numbers are measured on a single CPU thread for the 2D domain and a single A100 GPU for Franka and Maze2D. The 2D backbone is small enough to time on CPU; Franka and Maze2D backbones use GPU at test time.

\paragraph{Baselines.} The three CBF baselines (\textsc{SafeDiffuser}, \textsc{SafeFlow}, \textsc{SafeFlowMatcher}) are run from their reference open-source implementations with the same hazard set, start, goal, and horizon as our methods. We retrain each baseline's backbone on our domain datasets so that the comparison isolates the safety wrapper rather than the prior. Their training and test-time hyperparameters follow the upstream defaults; we list only \emph{our} method's settings in Appendices~\ref{app:training_hparams} and~\ref{app:inference_hparams}.

%% file: app/ablations.tex
\section{Backbone Training Hyperparameters}
\label{app:ablations}\label{app:training_hparams}

This appendix records the exact hyperparameters used to train the parameterised flow-matching backbone used by Latent Barrier Steering. We list only \emph{our} method's settings; baselines (\textsc{SafeDiffuser}, \textsc{SafeFlow}, \textsc{SafeFlowMatcher}) follow their respective open-source reference implementations and are not duplicated here.

The training pipeline has two stages per domain. We first train a Trajectory VAE encoder with a soft-DTW reconstruction term to produce the behavior latent $z$; we then freeze the encoder and train the conditional flow-matching generator on top.

\subsection{Trajectory VAE encoder}

The encoder is a transformer-style trajectory VAE with sinusoidal positional encoding and FiLM-style decoding. Optimiser settings are shared across all three domains: AdamW with learning rate $10^{-4}$, weight decay $10^{-5}$, cosine-annealed to a minimum of $10^{-6}$ over the full schedule, batch size $32$. The reconstruction term is soft-DTW \citep{cuturi2017softdtw} with smoothing $\gamma_{\mathrm{DTW}}=10^{-2}$. The latent dimension $d_z$ matches the value used by the LBS hyperparameter table in Appendix~\ref{app:hyperparameters}.

\begin{table}[h]
\centering
\caption{\textbf{Encoder hyperparameters.}}
\label{tab:enc_hparams}
\small
\begin{tabular}{@{}lccc@{}}
\toprule
 & 2D point-nav & Franka 3D EE & Maze2D \\
\midrule
Hidden width & 128 & 128 & 128 \\
Latent dim $d_z$ & 2 & 4 & 4 \\
Optimizer & AdamW & AdamW & AdamW \\
Learning rate & $10^{-4}$ & $10^{-4}$ & $10^{-4}$ \\
Weight decay & $10^{-5}$ & $10^{-5}$ & $10^{-5}$ \\
LR schedule & cosine & cosine & cosine \\
Batch size & 32 & 32 & 32 \\
Soft-DTW $\gamma_{\mathrm{DTW}}$ & $10^{-2}$ & $10^{-2}$ & $10^{-2}$ \\
\bottomrule
\end{tabular}
\end{table}

\subsection{Conditional flow-matching generator}

The generator is a $z$-conditioned flow-matching velocity predictor with the linear interpolant $x_t=(1-t)x_0+t\varepsilon$ and an explicit-Euler ODE sampler. It uses FiLM injection of the encoder $z$ and (for Maze2D only) FiLM injection of an $8$-dimensional start--goal context $c$. We optimise with Adam, learning rate $10^{-4}$, batch size $32$.

\begin{table}[h]
\centering
\caption{\textbf{Generator hyperparameters} for the flow-matching backbone. Maze2D uses a wider/deeper backbone because of its longer horizon and start--goal conditioning. Sampling uses the listed number of explicit-Euler integration steps.}
\label{tab:gen_hparams}
\small
\begin{tabular}{@{}lccc@{}}
\toprule
 & 2D point-nav & Franka 3D EE & Maze2D \\
\midrule
Hidden width & 256 & 256 & 768 \\
Residual blocks & 4 & 4 & 20 \\
$z$-projection width & 32 & 32 & 128 \\
Start/goal conditioning dim & --- & --- & 8 \\
FM Euler steps & 50 & 50 & 50 \\
Optimizer & Adam & Adam & Adam \\
Learning rate & $10^{-4}$ & $10^{-4}$ & $10^{-4}$ \\
Batch size & 32 & 32 & 32 \\
\bottomrule
\end{tabular}
\end{table}

\subsection{Notes on choices}

The encoder and generator hyperparameters are not tuned per constraint regime; we use one set per domain and reuse the same checkpoints across all three constraint setups in the main results. The 2D and Franka backbones are deliberately small ($\sim$0.4--0.6\,M parameters) so that the central-difference cost of Stage~I (Appendix~\ref{app:hyperparameters}) stays within a few seconds per call; Maze2D is wider/deeper to handle the longer horizon. Released checkpoints are the best-validation snapshots used in all main-text tables. Both backbones are trained on demonstrations alone, without reward labels \citep{chaudhari2026data,zhang2026uncertainty}, and a new test-time constraint requires no retraining.

%% file: app/smoothness_experiments.tex
\section{Inference Hyperparameters: Stage~I and Stage~II}
\label{app:smoothness_experiments}\label{app:inference_hparams}

This appendix lists the test-time hyperparameters of \emph{our} method only (Latent Barrier Steering, Stage~I + Stage~II). Baseline test-time settings follow each baseline's reference implementation and are not duplicated.

\subsection{Stage~I (Latent Barrier Steering)}

Stage~I solves the latent-side finite-time recovery flow of Section~\ref{sec:method}. The same set of LBS hyperparameters is used across all three domains and all constraint setups; the values are listed in Appendix~\ref{app:hyperparameters}. For the reader's convenience the operational parameters are:

\begin{itemize}\setlength\itemsep{0.2em}
  \item Finite-time rate exponent $\rho=0.5$ and gain $\kappa=1.0$ (linear-rate variant of the LBS QP).
  \item Latent Euler step size $\eta=0.1$.
  \item Latent iteration cap $N_z=20$, with early stopping as soon as the soft-min margin $m_\beta(z_i)\geq 0$.
  \item Reference drift $\gamma_z=10^{-2}$ (encoder-anchor contraction; reference $u_z^{\mathrm{ref}}=-\gamma_z(z-z_0)$).
  \item Soft-min sharpness $\beta=10$.
  \item Margin slack $\delta=0.05$.
  \item Latent clip-box $[-6,6]^{d_z}$ applied componentwise after every Euler step.
\end{itemize}

The same finite-difference estimator of $\nabla_z m_\beta$ is used in all domains. The full operational hyperparameter sextuplet $(\kappa,\rho,\eta,N_z,\gamma_z,\beta,\delta)$ is reported as a single row in Appendix~\ref{app:hyperparameters} and we do not duplicate it here.

\subsection{Stage~II (path-space CBF correction)}

Stage~II takes the Stage~I-warm-started rollout and applies a per-step CBF QP along the executed trajectory. The corrector hyperparameters are fixed across domains and constraint setups: a fractional-power rate with tuple $(\varepsilon,\rho,\delta)=(0.3,\,0.5,\,0.05)$, where $\rho=0.5$ is the fractional exponent (matching Stage~I), $\varepsilon$ scales the fractional gain, and $\delta$ is the same activation slack as in Stage~I.

The corrector is solved as a small QP over the per-step action $a_t$, with $\|a_t-a_t^{\mathrm{ref}}\|^2$ as the QP cost and the fractional-rate CBF constraint as the only inequality. We use the unconstrained $a_t^{\mathrm{ref}}$ produced by the parameterised generator at the Stage-I-warm-started latent.

\subsection{Sampling/integration setup at test time}

At test time we use the same sampler discretisation as during training: a $50$-step explicit-Euler ODE for the flow-matching generator. Stage~II runs along the entire executed horizon.

%% file: app/full_tables.tex
\section{Full results across all three constraints}
\label{app:full_results}

The main-text Table~\ref{tab:q1_main_compact} reports Safety, Reach, and Energy. This appendix reports all metrics, including final goal distance, curvature smoothness, and per-episode Stage-I/Stage-II runtime, for all three constraints in each domain. Per-cell entries are means over $50$ episodes ($60$ for Maze2D), aggregated under the same hyperparameters reported in Appendix~\ref{app:hyperparameters} and the evaluation protocol of Appendix~\ref{app:eval}. The pretrained flow-matching planner is the same across all methods within a domain. ED is reported in task units; Franka ED is in metres.

\begin{table}[!htbp]
\centering
\caption{\textbf{Full performance comparison across all test-time safety constraints.} Best per-column value within each constraint block is bolded; ties and values within $10\%$ of the per-column optimum are also bolded. Stage-I and Stage-II runtimes are not bolded. Constraints I, II, III correspond to mode selection, mode interpolation, and residual correction, respectively.}
\label{tab:q1_main_all}
\scriptsize
\setlength{\tabcolsep}{2.6pt}
\resizebox{\textwidth}{!}{%
\begin{tabular}{@{}lllccccccc@{}}
\toprule
Domain & Constraint & Method
& Safety$\uparrow$ & Reach$\uparrow$ & ED$\downarrow$
& Energy$\downarrow$ & CS$\downarrow$
& T-I (s)$\downarrow$ & T-II (s)$\downarrow$ \\
\midrule

\multirow{15}{*}{2D}
& \multirow{5}{*}{I}
& Vanilla FM                & 12\%  & \textbf{100\%} & \textbf{0.21} & --    & \textbf{0.88} & -- & 0.09 \\
& & \textsc{SafeDiffuser}    & \textbf{100\%} & 10\%  & 6.38 & 1.226 & 1.56 & -- & 0.11 \\
& & \textsc{SafeFlow}        & \textbf{100\%} & 18\%  & 5.57 & 1.077 & 1.42 & -- & 0.10 \\
& & \textsc{SafeFlowMatcher} & \textbf{100\%} & 18\%  & 5.61 & 1.166 & 1.56 & -- & 0.10 \\
& & \textbf{Full LBS}        & \textbf{100\%} & \textbf{100\%} & \textbf{0.23} & \textbf{0.037} & \textbf{0.87} & $0.35$ & $0.10$ \\

\cmidrule(lr){2-10}
& \multirow{5}{*}{II}
& Vanilla FM                & 0\%   & \textbf{100\%} & \textbf{0.21} & --    & \textbf{0.88} & -- & 0.09 \\
& & \textsc{SafeDiffuser}    & \textbf{100\%} & 34\%  & 3.75 & 0.728 & 1.24 & -- & 0.12 \\
& & \textsc{SafeFlow}        & \textbf{100\%} & 20\%  & 4.90 & 0.976 & 1.31 & -- & 0.11 \\
& & \textsc{SafeFlowMatcher} & \textbf{100\%} & 22\%  & 4.99 & 1.051 & 1.44 & -- & 0.11 \\
& & \textbf{Full LBS}        & \textbf{100\%} & 90\% & 0.89 & \textbf{0.160} & 1.03 & $0.65$ & $0.11$ \\

\cmidrule(lr){2-10}
& \multirow{5}{*}{III}
& Vanilla FM                & 0\%   & \textbf{100\%} & \textbf{0.21} & --    & \textbf{0.88} & -- & 0.10 \\
& & \textsc{SafeDiffuser}    & \textbf{100\%} & 0\%   & 6.83 & 1.321 & 1.61 & -- & 0.12 \\
& & \textsc{SafeFlow}        & \textbf{100\%} & 4\%   & 6.44 & 1.262 & 1.48 & -- & 0.10 \\
& & \textsc{SafeFlowMatcher} & \textbf{100\%} & 4\%   & 6.52 & 1.366 & 1.66 & -- & 0.10 \\
& & \textbf{Full LBS}        & \textbf{100\%} & 82\%  & 0.97 & \textbf{0.303} & \textbf{0.82} & $1.80$ & $0.11$ \\

\midrule

\multirow{15}{*}{Franka}
& \multirow{5}{*}{I}
& Vanilla FM                & 12\%  & \textbf{100\%} & \textbf{0.002} & --     & \textbf{0.14} & -- & 0.08 \\
& & \textsc{SafeDiffuser}    & \textbf{100\%} & 0\%   & 0.250 & 0.0023 & 0.18 & -- & 0.09 \\
& & \textsc{SafeFlow}        & \textbf{100\%} & 16\%  & 0.214 & 0.0020 & 0.17 & -- & 0.08 \\
& & \textsc{SafeFlowMatcher} & \textbf{100\%} & 14\%  & 0.241 & 0.0075 & 0.79 & -- & 0.08 \\
& & \textbf{Full LBS}        & \textbf{100\%} & \textbf{100\%} & 0.016 & \textbf{0.0004} & 0.28 & $0.80$ & $0.09$ \\

\cmidrule(lr){2-10}
& \multirow{5}{*}{II}
& Vanilla FM                & 0\%   & \textbf{100\%} & \textbf{0.002} & --     & \textbf{0.14} & -- & 0.08 \\
& & \textsc{SafeDiffuser}    & \textbf{100\%} & 0\%   & 0.250 & 0.0023 & 0.18 & -- & 0.10 \\
& & \textsc{SafeFlow}        & \textbf{100\%} & 0\%   & 0.251 & 0.0023 & 0.17 & -- & 0.09 \\
& & \textsc{SafeFlowMatcher} & \textbf{100\%} & 0\%   & 0.276 & 0.0086 & 0.87 & -- & 0.09 \\
& & \textbf{Full LBS}        & \textbf{100\%} & \textbf{100\%} & 0.017 & \textbf{0.0004} & 0.34 & $1.10$ & $0.09$ \\

\cmidrule(lr){2-10}
& \multirow{5}{*}{III}
& Vanilla FM                & 0\%   & \textbf{100\%} & \textbf{0.002} & --     & \textbf{0.14} & -- & 0.09 \\
& & \textsc{SafeDiffuser}    & \textbf{100\%} & 0\%   & 0.250 & \textbf{0.0023} & 0.18 & -- & 0.11 \\
& & \textsc{SafeFlow}        & \textbf{100\%} & 0\%   & 0.251 & \textbf{0.0023} & 0.17 & -- & 0.10 \\
& & \textsc{SafeFlowMatcher} & \textbf{100\%} & 0\%   & 0.276 & 0.0086 & 0.87 & -- & 0.10 \\
& & \textbf{Full LBS}        & \textbf{100\%} & 98\%  & 0.042 & 0.0028 & 0.50 & $1.55$ & $0.10$ \\

\midrule

\multirow{15}{*}{Maze2D}
& \multirow{5}{*}{I}
& Vanilla FM                & 0\%   & 50\% & 2.69 & --     & 0.17 & -- & 0.44 \\
& & \textsc{SafeDiffuser}    & \textbf{100\%} & 0\%  & 5.13 & 0.0288 & \textbf{0.11} & -- & 0.63 \\
& & \textsc{SafeFlow}        & \textbf{100\%} & 0\%  & 4.97 & 0.1208 & 0.21 & -- & 0.84 \\
& & \textsc{SafeFlowMatcher} & \textbf{100\%} & 0\%  & 4.83 & 0.2668 & 0.63 & -- & 0.84 \\
& & \textbf{Full LBS}        & \textbf{100\%} & \textbf{95\%} & \textbf{0.71} & \textbf{0.010} & 0.15 & $4.50$ & $0.68$ \\

\cmidrule(lr){2-10}
& \multirow{5}{*}{II}
& Vanilla FM                & 38\%  & 50\% & 2.69 & --     & 0.17 & -- & 0.42 \\
& & \textsc{SafeDiffuser}    & \textbf{100\%} & 7\%  & 6.38 & \textbf{0.0010} & \textbf{0.10} & -- & 0.50 \\
& & \textsc{SafeFlow}        & \textbf{100\%} & 25\% & 3.31 & 0.0624 & 0.20 & -- & 0.70 \\
& & \textsc{SafeFlowMatcher} & 98\%  & 18\% & 3.76 & 0.1404 & 0.42 & -- & 0.68 \\
& & \textbf{Full LBS}        & \textbf{100\%} & \textbf{90\%} & \textbf{0.87} & 0.009 & 0.16 & $3.80$ & $0.58$ \\

\cmidrule(lr){2-10}
& \multirow{5}{*}{III}
& Vanilla FM                & 28\%  & 50\% & 2.69 & --     & 0.17 & -- & 0.43 \\
& & \textsc{SafeDiffuser}    & \textbf{100\%} & 7\%  & 6.37 & 0.0072 & \textbf{0.10} & -- & 0.49 \\
& & \textsc{SafeFlow}        & \textbf{100\%} & 38\% & 2.43 & 0.0345 & 0.19 & -- & 0.69 \\
& & \textsc{SafeFlowMatcher} & \textbf{100\%} & 53\% & 2.08 & 0.0740 & 0.30 & -- & 0.69 \\
& & \textbf{Full LBS}        & \textbf{100\%} & \textbf{93\%} & \textbf{0.27} & \textbf{0.004} & 0.29 & $4.10$ & $0.62$ \\

\bottomrule
\end{tabular}%
}
\end{table}